%% file: main.tex
\documentclass[letterpaper,journal]{IEEEtran}
\usepackage{amsmath,amsfonts}
\usepackage{algorithmic}
\usepackage{algorithm}
\usepackage{array}
\usepackage[caption=false,font=footnotesize,labelfont=rm,textfont=rm]{subfig}
\usepackage{textcomp}
\usepackage{stfloats}
\usepackage{url}
\usepackage{verbatim}
\usepackage{graphicx}
\usepackage{cite}
\usepackage{ragged2e}
\usepackage{tabularx}
\usepackage{threeparttable}
\usepackage{booktabs}
\usepackage{multirow}
\usepackage{color}
\usepackage{makecell}
\usepackage{colortbl}
\usepackage{xcolor}

\begin{document}

% \title{Mapping-MVP: Mapping in Monocular, Vectorized, \\ and Probabilistic Way for High-definition Map Generation}

\title{Enhancing Inverse Perspective Mapping for Automatic Vectorized Road Map Generation}

% Anonymized
% \author{}
\author{Hongji Liu, Linwei Zheng, Yongjian Li, Mingkai Tang, Xiaoyang Yan, Ming Liu, \textit{Fellow IEEE}, and Jun Ma 
% \textit{Senior Member, IEEE}
        % <-this % stops a space
\thanks{H. Liu, M. Tang, and X. Yan are with The Hong Kong University of Science and Technology, Hong Kong SAR, China (e-mail: hliucq@connect.ust.hk).
}% <-this % stops a space
\thanks{L. Zheng, Y. Li, and M. Liu are with The Hong Kong University of Science and Technology (Guangzhou), Guangzhou 511453, China (e-mail: lzhengad@connect.hkust-gz.edu.cn).
}% <-this % stops a space
\thanks{J. Ma is with the Robotics and Autonomous Systems Thrust, The Hong Kong University of Science and Technology (Guangzhou), Guangzhou 511453, China, and also with the Division of Emerging Interdisciplinary Areas, The Hong Kong University of Science and Technology, Hong Kong SAR, China (e-mail: jun.ma@ust.hk).} 
}



\maketitle

\begin{abstract}
In this study, we present a low-cost and unified framework for vectorized road mapping leveraging enhanced inverse perspective mapping (IPM). In this framework, Catmull-Rom splines are utilized to characterize lane lines, and all the other ground markings are depicted using polygons uniformly. The results from instance segmentation serve as references to refine the three-dimensional position of spline control points and polygon corner points. In conjunction with this process, the homography matrix of IPM and vehicle poses are optimized simultaneously. Our proposed framework significantly reduces the mapping errors associated with IPM. It also improves the accuracy of the initial IPM homography matrix and the predicted vehicle poses. Furthermore, it addresses the limitations imposed by the coplanarity assumption in IPM. These enhancements enable IPM to be effectively applied to vectorized road mapping, which serves a cost-effective solution with enhanced accuracy. In addition, our framework generalizes road map elements to include all common ground markings and lane lines. The proposed framework is evaluated in two different practical scenarios, and the test results show that our method can automatically generate high-precision maps with near-centimeter-level accuracy. Importantly, the optimized IPM matrix achieves an accuracy comparable to that of manual calibration, while the accuracy of vehicle poses is also significantly improved.
\\

\textit{Note to Practitioners}—Creating accurate, high-definition road maps for autonomous vehicles is essential but often expensive and slow. Our work offers a practical, low-cost solution using just a standard front-facing camera. We automatically generate detailed vector maps—showing lane lines, crosswalks, arrows, and other markings—with near-centimeter precision. The key innovation is that our system learns and refines the camera's mapping to the road in real-time as the vehicle drives, correcting errors from movement and slight changes in camera position. This eliminates the need for time-consuming, individual calibration of each vehicle, making map creation much faster, cheaper, and scalable for large fleets. While it works best with clear road markings, our method currently relies on visible, well-defined markings and may struggle in conditions with poor visibility, heavy rain, or faded markings. Our method can be extended to other environments and even updated continuously using data from many vehicles, making it ideal for real-world deployment. Beyond autonomous driving, this technology is valuable for creating and dynamically maintaining environmental information for digital twins or automated mobile robots, enabling them to navigate and interact with their surroundings more intelligently and safely.
\end{abstract}


\input{sections/intro}
\input{sections/related_work}

\input{sections/methodology}

\input{sections/exp}
\input{sections/conclusion}

\bibliographystyle{IEEEtran}
\bibliography{IEEEabrv,ref}

% \newpage

 

% \end{IEEEbiography}



\vfill

\end{document}

%% file: sections/intro.tex
\section{Introduction\label{sec: intro}}
\subsection{Motivation}
%interesting
\IEEEPARstart{R}{ecent} developments in simultaneous localization and mapping (SLAM) technology have rendered it possible to produce metric maps for unmanned ground vehicles (UGVs)~\cite{Zhang-RSS-14, legoloam2018, ye2019tightly, Jiao_2024}. However, creating these maps for large-scale open-road environments remains challenging. The significant storage and computational demands for handling the extensive data volume of the metric maps increase the cost of UGVs, restricting their widespread application~\cite{cadena2016past}. Recently, there has been a trend towards the use of vectorized high-definition (VHD) maps, which are continually gaining traction in numerous fields~\cite{10018201, 8468109, 10018862, 10818985, 10786263, 9304783, 9827297, 9636746} and have sparked a considerable amount of related research~\cite{10508471, 8946549}. VHD maps describe traffic elements through vector data such as points, curves, and polygons with geographical location and associated semantic information. Key advantages of VHD maps include their lightweight nature, rich attribute information, and simplicity in depicting road networks. Due to these benefits, VHD maps present a viable alternative to traditional metric maps in the UGV contexts~\cite{10497895, 10336514,10184094}. Table~\ref{tab: vhdmapvspointcloud} offers a comparative analysis of VHD maps and point cloud maps (a classic type of metric map), highlighting their respective advantages and disadvantages from various perspectives.

Accurate coordinates of ground markings are essential for VHD maps. However, generating these maps often relies heavily on manual surveying, which is labor-intensive and time-consuming, particularly in large-scale scenarios. Additionally, any stain or change on the markings necessitates re-measurement or update of the map. Therefore, developing a low-cost, efficient automated system for VHD map generation would greatly improve the practicality and scalability of their applications. 

\begin{figure}[!t] 
    \centering
    \includegraphics[width=0.48\textwidth]{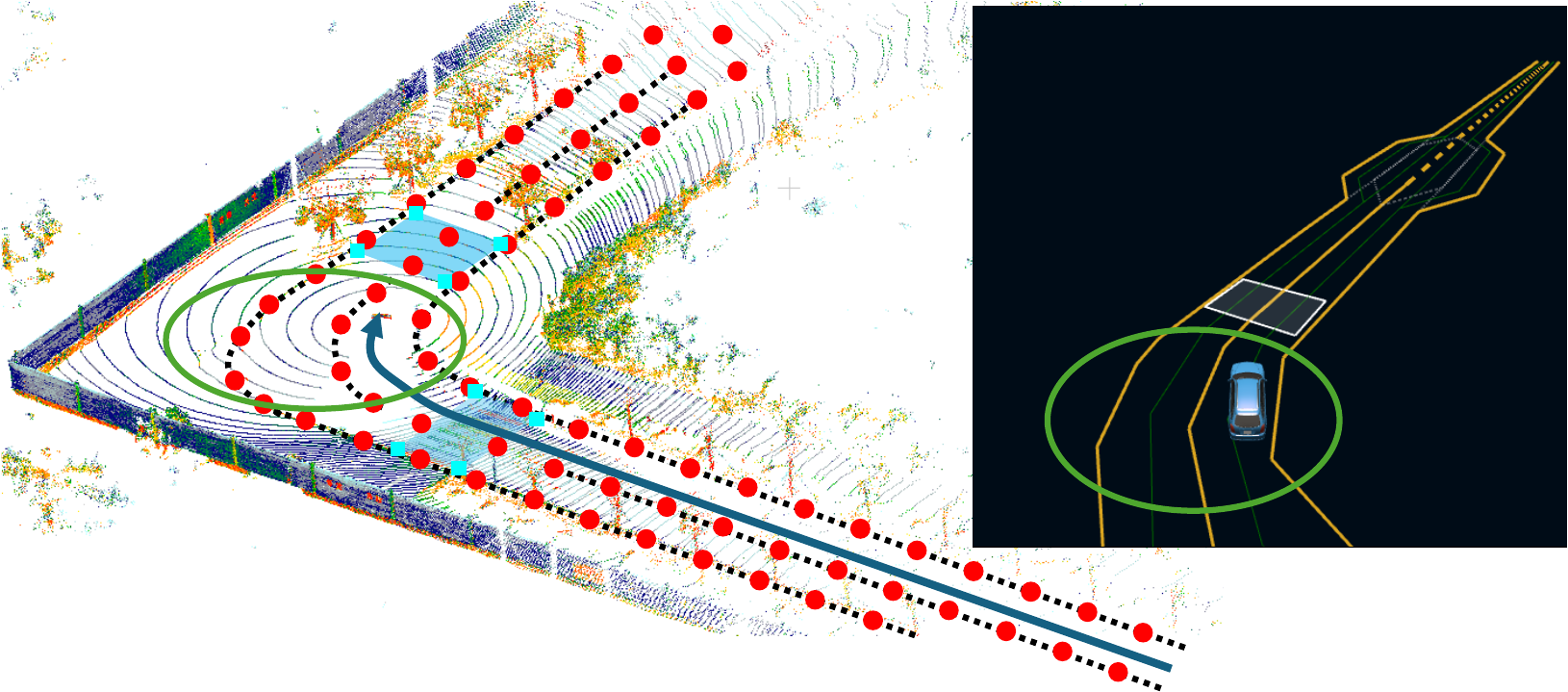}
    \caption{An example of the generated vectorized HD map. At the turn, there is a pedestrian crossing at both the front and rear of the UGV, which are marked by blue polygons. The corner points of them are stored in the map and marked in cyan in the point cloud. The lane lines are generated as Catmull-Rom splines. The control points of them are shown as red spheres in the point cloud, the sampled lane line points are marked as black points.
    }
    \label{fig: mf}
\end{figure}

Inverse perspective mapping (IPM) presents a promising, cost-effective approach for automatic vectorized mapping. It transforms forward-facing camera images into bird's-eye view (BEV) images using a homography matrix, also known as the IPM matrix. Remarkably, even a single monocular camera can accomplish this task in simple scenarios. Moreover, the IPM process is mathematically traceable, allowing for better modeling of map uncertainties. This transparency offers an advantage over deep learning models, which often function as black boxes when predicting ground markings.

% In addition, the spatial uncertainties of the road markings are crucial for some downstream tasks. Taking UGV localization as an example, in VHDMap-SE~\cite{10497895}, the authors cannot obtain the uncertainty of VHD map, and can only determine the map uncertainty through the verification step (calculating the deviation between the map lane width and the actual value). These uncertainties indicate the confidence level of the reference information provided by the map, and can also provide guidance for the optimization focus of the map. The uncertainty of maps generated based on different methods varies, so it is necessary to model the uncertainty of maps while building them.

%hard

\subsection{Challenges}
% ~\cite{hartley2003multiple}
% \IEEEpubidadjcol
Despite its utility, establishing VHD maps through IPM presents several challenges. Calibrating the IPM matrix for large-scale UGV applications is both time-consuming and labor-intensive. For instance, in port applications, UGVs are often produced in large numbers, with fleets typically comprising hundreds of units. Each UGV requires a unique IPM matrix, necessitating individual calibration~\cite{yu2022accurate,10588820}. Additionally, the accuracy of corner coordinates of markings derived via IPM diminishes when the vehicle is in motion~\cite{jeong2016adaptive}. This inaccuracy arises for two main reasons. First, the planar assumption underpinning IPM is not always valid. Variations in road conditions can result in the calibration plane differing from the actual road surface. Second, slight changes in the camera's position can alter the calibrated IPM matrix. Moreover, IPM assumes a flat, single-plane ground, preventing it from estimating the height of ground markings~\cite{roadmap}.

\begin{table}[!t]
    \centering
    \caption{Comparison of VHD maps and Pointcloud Maps. The bold text indicates the one with better performance. The reasons for the evaluation are briefly written in parentheses.}
    \renewcommand\arraystretch{1.2} % adjust line spacing
    \begin{threeparttable}
        \begin{tabularx}{0.95\linewidth}{>{\centering\arraybackslash}m{15mm}|>{\raggedright\arraybackslash}m{30mm}|>{\raggedright\arraybackslash}m{30mm}}
        \hline
        Comparison Items & \multicolumn{1}{c|}{VHD Map} & \multicolumn{1}{c}{Point Cloud Map} \\
        \hline
        Map Storage  & \textbf{Low (The map is organized using vectorized data)} & High (The map stores all original points)\\
        \hline
        Semantic Level & \textbf{High (The map includes semantic labels)} & Low (The map only contains raw data)  \\
        \hline
        % \rowcolor[HTML]{E7E6E6}
        Construction Difficulty & High (The construction requires human interaction with low-level automation) & \textbf{Low (The pipeline is automated by SLAM technologies)}  \\
        \hline
        Maintenance Difficulty & \textbf{Mid (The update and modification are based on elements)} & High (The update and modification require partial map reconstruction and map stitch) \\
        \hline
        % \rowcolor[HTML]{E7E6E6}
        Map Precision & High/Low (The precision depends on the degree of automation) & High/Low (The precision depends on scenario, algorithm, and sensor setup) \\
        % \rowcolor[HTML]{E7E6E6}
        % \;Ours & $\checkmark$ & HDF; & CA; & I-R \\
        \hline
        \end{tabularx}
        % \smallskip
        % \scriptsize
        % \begin{tablenotes}
        % \RaggedRight
        % \item[*] The dark-colored items indicate the parts of the VHD map that need to be improved.\\
        % \end{tablenotes}
        \label{tab: vhdmapvspointcloud}
    \end{threeparttable}
\end{table}
Beyond the challenges inherent to the IPM method, the representation of strip-shaped markings (including lane lines and road boundaries) is also evolving~\cite{homayounfar2019dagmapper, xu2021topo, tang2022thmatencenthdmap}. Spline curves are increasingly used for lane lines to better handle curved scenes and reduce storage needs. However, this approach introduces more complex optimization requirements. In~\cite{qiao2023online}, the authors proposed to optimize the position of spline control points according to the position of predicted 3D lane line points. Unfortunately, similar methods are unsuitable for IPM. This is because the predicted 3D positions of lane markings by IPM contain errors, making it unreasonable to use this data for optimizing control points. In addition, how to unify the expression of ground markings is also a challenge. Most researchers overlook the issue of their concise expression and directly use the results of instance segmentation or LiDAR intensity~\cite{9636419, 9636227, jang2021lane,9807400}. Some researchers choose to use corner points to express ground markings because they are easy to extract and can accurately represent the type and location of the markers~\cite{ranganathan2013light, cheng2021road, 10588820}. However, this method can only be applied to markings with corners, those markings without obvious corners, such as speed limit markings, cannot be expressed. In addition, there are some markings with a large number of corners that are complicated to extract completely, such as zebra crossings. Considering the universality of the expression, it is required to take all types of ground markings into account, rather than just a few that are more distinctive.

\subsection{Contributions}
%solvable
To tackle the aforementioned challenges, we propose a vectorized road mapping system that enhances IPM. The system initially transforms the polygon corners of the detected road markings and lane points from images into the vehicle frame using IPM. Subsequently, these points are converted into the map frame based on the vehicle's pose, while simultaneously estimating their uncertainty. To tackle the issue of representations for strip-shaped markings, we employ Catmull-Rom splines to model lane lines. We infer the position of control points using lane points derived from IPM, considering the uncertainty estimates to exclude points with high uncertainty. This process enhances the accuracy of control point estimation. In order to express ground markings in a unified form, regardless of their shape and type, we use their bounding boxes (polygons) to represent them. In this way, only the position of the corners of their bounding boxes is required to store them.

To further refine the mapping accuracy and address calibration challenges, we implement a sequential optimization of the map and vehicle poses. During map optimization, we also refine the IPM matrix to improve its precision, through which to replace the calibration process. The optimization objective ensures that projected markings on the image plane align closely with segmentation results. This approach significantly enhances the accuracy of maps generated through IPM. Additionally, it simplifies the calibration process of IPM matrices for large UGV fleets, eliminating the need for individual calibration. A rough initial estimation is sufficient, as each UGV's IPM matrix is automatically fine-tuned during operation. These refined matrices can then be applied to other UGV tasks~\cite{oliveira2015multimodal, 8814056, pan2020cross}.

We summarize the primary contributions of this work as:
\begin{enumerate}
    \item We designed a feasible and unified low-cost vectorized mapping framework for generating road maps automatically based on IPM. In the framework, lane lines are modeled using the Catmull-Rom spline, and other types of ground markings are represented by a polygon (bounding box) uniformly.
    \item We proposed a method for optimizing the 3D coordinates of ground markings and Catmull-Rom spline control points guided by instance segmentation. Through the method, the IPM homography matrix and vehicle poses could be jointly optimized with the map points. 
    \item We performed an extensive evaluation of the system in two practical scenarios, including automated ports and a public road. The optimized vehicle poses, IPM matrix, and vectorized map precision are evaluated in detail. The results showed that our method could automatically generate high-precision maps with near-centimeter-level accuracy. In addition, the optimized IPM matrix reached an accuracy similar to that of manual calibration and the pose accuracy is improved effectively.
\end{enumerate}

%% file: sections/related_work.tex
\section{Related Work\label{sec: related_work}}
In recent years, the demand for road maps by UGVs has evolved. The focus has shifted from pursuing integrity and accuracy to emphasizing local topological structure~\cite{10670223,10321736}. Unlike original HD maps, these local maps do not require strict accuracy for traffic elements. Methods proposed in~\cite{lu2019monocular, roddick2020predicting, can2022understanding, 9811901, yang2021projecting}
% mani2020monolayout
used monocular camera images for semantic map creation via PV-to-BEV (perspective view to BEV) transformation. Some approaches reconstructed comprehensive surrounding maps using multiple images. For instance, Pan \textit{et al.}~\cite{pan2020cross} developed the view parsing network to convert multi-angle observations into a BEV semantic map. Philion \textit{et al.}~\cite{philion2020lift} deduced road semantics directly in the BEV generated from arbitrary camera rigs. Deng \textit{et al.}~\cite{deng2019restricted} used multiple panoramic cameras for specialized BEV generation. HDMapNet~\cite{li2022hdmapnet} and VectorMapNet~\cite{pmlr-v202-liu23ax} introduced vectorization in BEV maps. MapTR~\cite{MapTR} further unified the shape modeling of map elements and solved the ambiguity problem in map element matching. However, these methods prioritized semantic details over accuracy, which posed challenges for downstream tasks requiring precise HD maps (e.g., localization). Moreover, due to BEV characteristics, these methods cannot estimate the three-dimensional coordinates of map elements. The changing demand for road maps does not eliminate the need for VHD maps. Instead, both types complement each other. For areas not covered by original HD maps, an online topology map is necessary for downstream tasks. In areas with VHD maps, UGVs prefer them for high-precision, beyond-line-of-sight perception, reducing the perception module's load. Thus, the accuracy of the map remains essential.

There has also been significant progress in high-precision road mapping in recent years. Elhousni \textit{et al.}~\cite{elhousni2020automatic} used inverse projection to map road surfaces, curbs, and lanes onto point clouds with calibrated camera-LiDAR transformations.  Similarly, Zhou \textit{et al.}~\cite{zhou2021automatic} inversely projected image segmentation results onto the ground plane derived from point cloud data. However, both approaches were costly as they rely on point clouds for real-world coordinates. They were impractical for low-cost UGVs lacking LiDAR due to constraints like cost and space. On the contrary, monocular cameras were already common in UGVs and offered a cost-effective alternative for HD map generation. Therefore, monocular cameras are adopted in our work to simplify mapping setup requirements.

Some researchers who shared similar considerations with us also used monocular cameras to estimate the three-dimensional positions of ground markings. PersFormer~\cite{10.1007/978-3-031-19839-7_32} is a cutting-edge method for 3D lane detection, predicting 3D lanes by establishing 3D lane anchors in BEV space. In~\cite{qiao2023online}, Catmull-Rom spline optimization was added to PersFormer, enhancing accuracy and robustness in lane mapping.  However, both of them cannot predict the location of other ground markings. Additionally, deep learning's black-box nature prevented estimating the uncertainty of map elements. In our work, we employed IPM for real-time mapping of all types of ground markings. The mathematical traceability of IPM~\cite{9850364} allowed us to estimate the uncertainty in the elements of the map. We used this uncertainty to filter out estimates with high uncertainty.

Similar to our approach, several methods also utilized monocular cameras with IPM for road element mapping. Guo~\textit{et al.}~\cite{guo2016low} generated BEV imagery and a road lane graph via homography transformation. Ranganathan \textit{et al.}~\cite{ranganathan2013light} mapped road marking corners using IPM. Jang \textit{et al.}~\cite{jang2018road} employed graph optimization to represent lane elements as nodes, constructing lane-level HD maps. In subsequent work~\cite{jang2021lane, cheng2021road}, road markings were also converted into graph nodes and optimized alongside the vehicle poses. Qin \textit{et al.}~\cite{roadmap} inversely projected the semantic elements from images into point clouds for local mapping. However, these methods focused on improving vehicle pose estimation for mapping accuracy, neglecting IPM's inherent errors. They also required precise calibration of the IPM matrix initially. In contrast, our earlier work PGO-IPM~\cite{10588820}, refined both the map and the preliminary IPM matrix, improving HD map quality and IPM matrix precision simultaneously. Yet, PGO-IPM did not address the coplanar constraint of IPM or predict the z-axis coordinates of markings. It also overlooked potential vehicle pose errors and lacked lane line modeling. In this work, we further added the $z$-axis coordinates to the optimization variables, and introduced a pose update process post-map optimization to reduce pose errors. Importantly, we extended PGO-IPM to include lane lines modeling, enhancing system completeness and generality.
% However, the IPM matrix on which these methods depend also possesses errors, so having better vehicle poses does not completely guarantee that a more accurate HD map can be established. 

% In addition, none of the above methods mentioned the data association problem. In a scenario with many similar and close markings, it is challenging to judge whether the markings observed in the current frame and the markings observed in the previous frames are the same. Therefore, there is a potential risk of erroneous data association in the subsequent optimization process.

% All the above methods considered the IPM and the construction of high-precision map as two independent tasks. We are the first to propose that these two independent tasks can be processed simultaneously.

%% file: sections/methodology.tex
\section{METHODOLOGY\label{sec: meth}}
\begin{figure*}[!t] 
    \centering
    \includegraphics[width=0.96\textwidth]{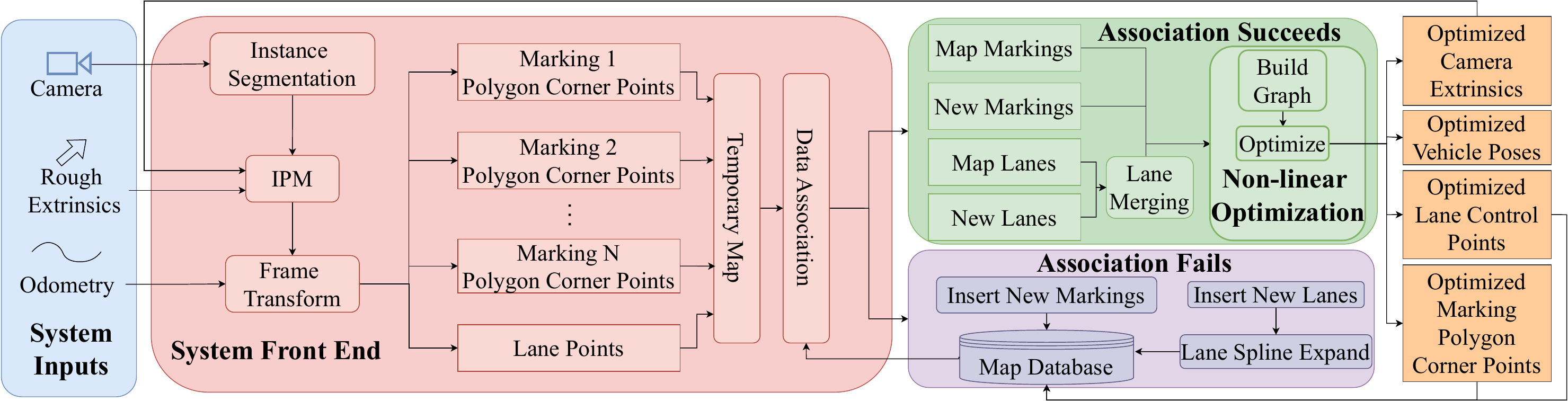}
    \caption{The workflow of our proposed system for automatic vectorized road mapping. The system can be divided into two parts. The front end is responsible for generating temporary maps and associating newly detected markings with those that have already existed on the map. When the data association succeeds, the back end jointly optimizes the map, camera extrinsics, and vehicle poses. The optimized map data updates the database, and the optimized camera extrinsic is fed forward to the front end.
    }
    \label{fig: system}
\end{figure*}
\subsection{Problem Statement}
\label{subsec: ps}
In our problem, the variables that need to be estimated include the position of map elements, IPM homography matrix, and vehicle poses, namely 
\begin{equation}
    \mathcal{X} = \{\mathbf{M}_{i\in [0, I]}^{m \in [0,4]}, \mathbf{L}_{j \in [0, J]}^{m \in [0, n]}, \mathbf{R}_{b}^c, \mathbf{t}_{b}^c, \mathbf{T}_{k \in [0, K]}\}
\end{equation}
where $\mathcal{X}$ is the set of all the variables, $\mathbf{M}_{i\in [0, I]}^{m \in [0,4]}$ represents the $m$th corner points of $i$th longitudinal markings' polygon, $\mathbf{L}_{j \in [0, J]}^{m \in [0, n]}$ represents the $m$th control points of $j$th lane's spline. $\mathbf{R}_b^c$ and $\mathbf{t}_b^c$ represent the rotation and translation transformations from the camera frame to the body frame, respectively. According to the derivation in~\cite{10588820}, the IPM homography matrix can be decomposed into the transformation relationship between the camera frame and the body frame. $\mathbf{T}_k$ represents the pose of the vehicle body frame to the world frame at timestamp $k$. We formulate the whole problem as a maximum likelihood estimation (MLE)~\cite{barfoot2017State} for all the measurements during the data collection process: 
\begin{equation}
    \hat{\mathcal{X}} = \underset{\mathcal{X}}{\operatorname {arg\,max}} \, P (\mathcal{Z}|\mathcal{X})  =\underset{\mathcal{X}}{\operatorname {arg\,min}} \, \sum_k f(\mathcal{X}_k,\mathbf{z}_k)
\label{eq: mle1}
\end{equation}
where $\mathcal{Z}$ is the set of all measurements $\mathbf{z}_k$ for the corner points of the bounding box of ground markings and the lane line points. Correspondingly, $\mathcal{X}_k$ represents the variables to be optimized corresponding to measurement $\mathbf{z}_k$. $f(\cdot)$ is the objective function. 
Assuming the measurement of target objects on the image plane follows the Gaussian distribution, \eqref{eq: mle1} can be solved as a non-linear least-squares problem:
\begin{equation}
    \hat{\mathcal{X}} = \underset{\mathcal{X}}{\operatorname {arg \, min}} = \sum_k {||\mathbf{r}(\mathcal{X}_k,\mathbf{z}_k)||}_\sigma^2 
    %= \argmin_{\mathcal{X}} \sum_k {||\mathbf{r}(\mathcal{X}_k,\mathbf{z}_k)||}_\sigma^2,
    \label{eq: mle2}
\end{equation}
where $\sigma$ is the covariance and $\mathbf{r}$ is the residual function. Such a problem can be solved using iterative methods such as Gauss-Newton or Levenberg-Marquardt. 

% \subsection{Road Marking Detection}

% For the lane detection task, we use UFLD~\cite{qin2020ultra} for its remarkable speed and accuracy. 
\begin{figure}[!t] 
    \centering
    \includegraphics[width=0.48\textwidth]{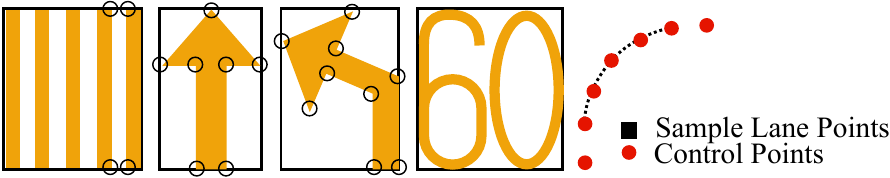}
    \caption{Different types of ground markings that possess various corner types. From left to right are zebra crossings, straight arrows, left turn arrows, speed limit signs, and lane lines. Each rectangle included in the zebra crossing contains four corners, which are not all marked in the figure. The speed limit signs do not include corners at all.
    }
    \label{fig: marking_type}
\end{figure}
\subsection{System Overview}
\label{subsec: so}
In general, the inputs of our vectorized road mapping system include any number of monocular camera images, the corresponding vehicle poses, and roughly estimated camera extrinsics. The markings detection module segments the markings and lanes in the image and extracts the key points. The map generation module constructs a temporary map using naive IPM followed by necessary frame transformation. The data association process determines whether a newly detected marking or lane has existed on the map. Every time new observation of the existing marking occurs (data association succeeds), the optimization module will optimize the position of the markings, the IPM homography matrix, and the vehicle poses jointly. The detailed composition of the system can be found in Fig.~\ref{fig: system}.

\subsection{Vectorized Representation of the Ground Markings}
We divide common ground markings into two categories: lateral markings and longitudinal markings. Lateral markings mainly include lane lines and road curbs that appear on the left and right sides of the vehicle's direction of travel. Longitudinal markings include arrows, pedestrian crossings, and other markings that appear on the front and rear sides of the vehicle when traveling.

\begin{figure}[!t] 
    \centering
    \subfloat[Diamond Markings Segmentation]{
            \includegraphics[width=0.47\textwidth]{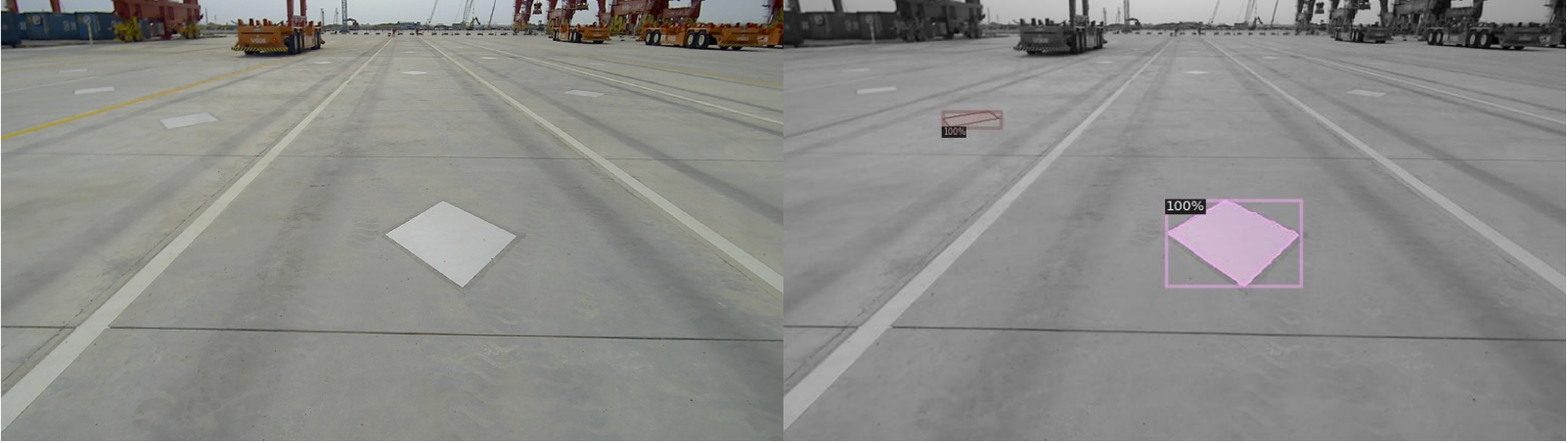}
            \label{subfig: diamond_seg}
        }  
    %\hfill
    \\
    \subfloat[Crosswalk Marking Segmentation]{
            \includegraphics[width=0.47\textwidth]{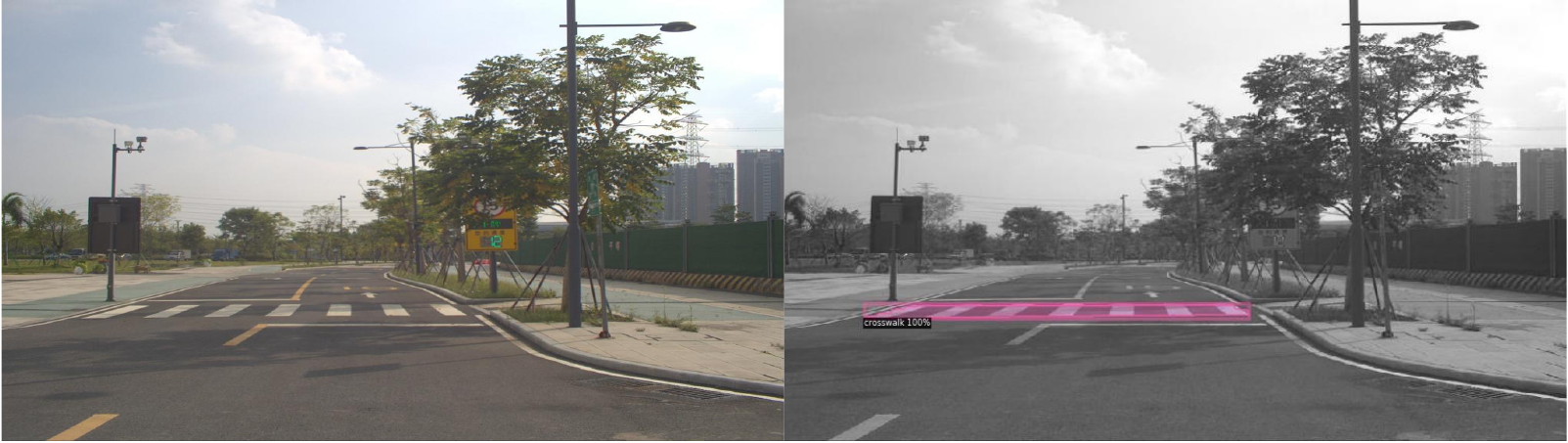}
            \label{subfig: crosswalk_seg}
        } \\
    \caption{Examples of ground markings segmentation. The type of markings can be self-defined in the system.}
    \label{fig: seg_example}
\end{figure}

In our previous work~\cite{10588820}, we expressed longitudinal markings by extracting the corners of the target markings. However, some ground markings do not feature clear corners or may possess intricate corners. Several examples of corner points for different types of ground markings are provided in Fig.~\ref{fig: marking_type}. Considering the convenience of instance segmentation and the universality of ground markings expression, we use the polygon outlined by the bounding box to represent different longitudinal ground markings as shown in Fig.~\ref{fig: seg_example}. This also brings the benefit of facilitating the matching of polygon corners during data association.

In~\cite{10588820}, the modeling of lane lines was achieved through line fitting to downsample lane line points. However, this approach cannot handle curved lane lines.  In~\cite{qiao2023online}, the authors introduced the benefits of using Catmull-Rom spline to represent lane lines. Inspired by these ideas, we use the Catmull-Rom spline to approximate the lane lines and optimize the control points to improve accuracy.

\subsection{IPM Error Guided Control Points Estimation}
Using Catmull-Rom splines to express lane lines involves estimating and optimizing the positions of control points. In~\cite{qiao2023online}, the authors introduced how to predict the position of control points based on estimated 3D lane markings. However, there is an estimation error in the position of lane points, which can lead to errors in the estimated position of control points. In addition, the lane line point estimation is based on deep learning models, and the uncertainty of each predicted point cannot be determined. For lane estimation based on IPM, the error of each lane marking point can be estimated. In~\cite{9850364}, the authors introduced a method for estimating IPM errors by considering the changes in camera height and pitch angle during the movement of vehicle, as well as the pixel errors generated during instance segmentation in the front-view image plane. The estimated IPM error of a point in the image plane can be calculated following

\begin{equation}
    \mathbf{\Omega}_{\mathbf{p}} = \mathbf{J}_{u,v}\mathbf{e}_{p}{\mathbf{e}^\top_{p}}{\mathbf{J}^\top_{u,v}} + \sigma_{\theta}^2\mathbf{J}_{\theta}{\mathbf{J}^\top_{\theta}} + \sigma_{h}^2\mathbf{J}_{h}{\mathbf{J}^\top_{h}},
\end{equation}
where $\mathbf{\Omega}_{\mathbf{p}}$ represents the uncertainty matrix of point $\mathbf{p}$ on IPM plane. $\mathbf{J}_{u,v}$ is the Jacobian of the final estimation error regarding pixel error vector $\mathbf{e}_{p}$. Similarly, $\mathbf{J}_{\theta}$ is the Jacobian regarding pitch angle estimation error $\sigma_{\theta}$, $\mathbf{J}_{h}$ is the Jacobian regarding camera height estimation error $\sigma_{h}$. Unlike in~\cite{9850364}, we model pixel errors as error vectors rather than the standard deviation of $u$-axis and $v$-axis errors. $\mathbf{\Omega}_{\mathbf{p}}$ is an uncertainty matrix rather than a specific value. The diagonal elements of the matrix represent the $x$, $y$, and $z$ coordinate variances of the point on the IPM plane. We take the trace of $\mathbf{\Omega}_{\mathbf{p}}$ as the uncertainty value of point $\mathbf{p}$, namely,
\begin{equation}
    \Omega({\mathbf{p}}) = \text{tr}(\mathbf{\Omega}_{\mathbf{p}}) = \sum_{i=1}^{n} {\mathbf{\Omega}_{\mathbf{p}}}_{ii}.
\end{equation}
We calculate $\Omega({\mathbf{p}})$ for all target points and then select the top $N$ points with the lowest uncertainty for control points prediction following 
\begin{equation}
    \begin{split}
         &{ \mathbf{C}_1, \mathbf{C}_2, \ldots, \mathbf{C}_M } = g(\tau(P)), \\
         \tau(P) &= \{ \mathbf{p}_1, \mathbf{p}_2, \ldots, \mathbf{p}_N \mid \mathbf{p}_i \in P \text{ and }  \\
         &\Omega({\mathbf{p}_1}) \leq \Omega({\mathbf{p}_2}) \leq \ldots \leq \Omega({\mathbf{p}_N}) \ldots \leq \Omega({\mathbf{p}_n})\},
    \end{split}
\end{equation}
where $\mathbf{C}_{*}$ represents the lane control points predicted by function $g$~\cite{qiao2023online}. $P$ represents the set of all the lane points derived from IPM.

% \subsection{Data Association}
% \begin{figure}[!t] 
%     \centering
%     \includegraphics[width=0.48\textwidth]{fig/theory/data_ass.pdf}
%     \caption{Marking Association Example.
%     }
%     \label{fig: data_ass}
% \end{figure}

\subsection{Map Optimization}
\begin{figure}[!t] 
    \centering
    \includegraphics[width = 0.475\textwidth]{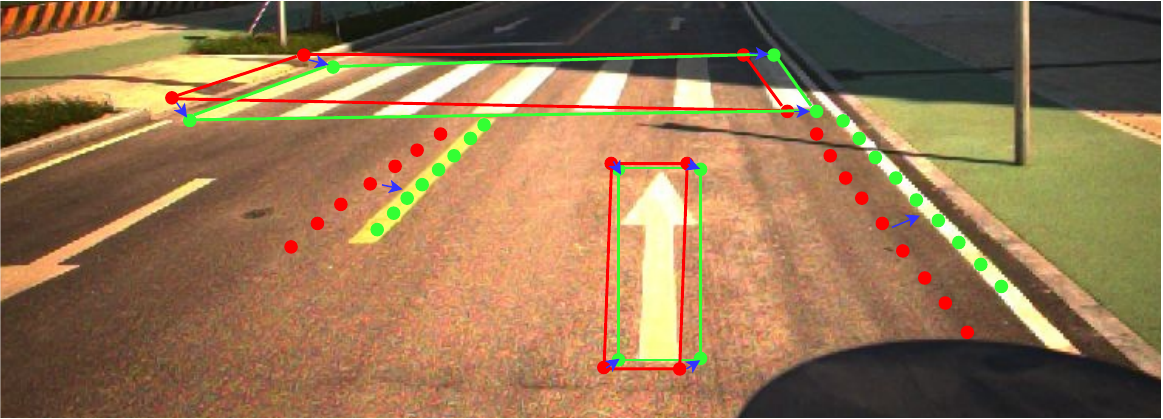}
    \caption{This is a visualization diagram of our general optimization objectives. The red markers in the figure represent the positions of map elements projected onto the camera plane from the generated map, while the green markers represent the results of instance segmentation. Our optimization goal is to reduce the offset between them (blue arrows).}
    \label{fig: optimization}
\end{figure}

To optimize the map, instead of directly fitting the contours\cite{jang2021lane} after projecting them to the ground plane, we optimize the homography matrix and the position of markings at the same time to minimize the displacement of the inversely projected markings to the observation. Fig.~\ref{fig: optimization} illustrates our objective of optimization.

\subsubsection{Longitudinal Markings}
For longitudinal markings, our optimization goal is to make the bounding box corners of the recognized target markings in the image match the projection results of their positions stored in the map. 
% The lack of rotation in UGV motion causes a degeneration of the translation part of the extrinsic matrix. Therefore, a translation prior (camera installation position)
% according to the installation drawings provided by the UGV manufacturer, 
% is used to constrain the extrinsic.
The initial guess of the extrinsic rotation matrix is based on assumptions about the vehicle body frame and camera frame. It assumes the $xy$-plane of the vehicle body frame is parallel to the ground, the $z$-axis is vertical to the ground and upward. The $z$-axis of the camera frame is parallel to the ground and forward, and the $y$-axis is vertical to the ground and downward.
% is provided as $z$ axis ahead and $x$ axis downward.
IPM provides the initial guess of each marking's position with a coarse IPM matrix at the beginning, then with a transformation composed of refined extrinsic and intrinsic after the first iteration. From this, we can further expand~\ref{eq: mle1} into
% \begin{equation}
% \begin{aligned}
% \hat{\mathcal{X}}&=\{\mathbf{\hat{R}}_b^c, \mathbf{\hat{t}}_{b}^c, \mathbf{\hat{M}}_i^m\} \\
% &=\underset{\mathbf{R}_{b}^c, \mathbf{t}_{b}^c, \mathbf{\hat{M}}_i^m}{\operatorname {arg\,min}} \, \Biggl\{ \sum_{i, m, k} f(i, m, k) ||\mathbf{\pi}_{ik}(\mathbf{M}_i^m) - \mathbf{z}^{im}_k||^2_\sigma \\ &\quad + ||\mathbf{t}_{b}^c - \mathbf{t}_0)||^2_\sigma \Biggl\},
% \end{aligned}
% \end{equation}
\begin{equation}
\begin{split}
\hat{\mathcal{X}}=\underset{\mathbf{R}_{b}^c, \mathbf{t}_{b}^c, \mathbf{M}_i^m}{\operatorname {arg\,min}} \, \Biggl\{ \sum_{i, m, k} f(i, m, k) \rho (||\mathbf{\pi}_{k}^{im}(\mathbf{M}_i^m) - \mathbf{z}^{im}_k||^2_\sigma) \\
+ ||\mathbf{t}_{b}^c - \mathbf{t}_0)||^2_\sigma \Biggl\},
\end{split}
\end{equation}
where $f(i,m,k)$ equals to 1 if $\mathbf{M}_i^m$ can be observed from the $k$th pose, otherwise it equals to 0. $\mathbf{\pi}_{k}^{im}$ is the function that projects $\mathbf{M}_i^m$ to the image plane according to the $k$th pose. $\mathbf{z}_{k}^{im}$ is the pixel measurement for $\mathbf{M}_i^m$ observed from $k$th pose. $\rho$ is the robust kernel function. The prior $\mathbf{t}_0$ is used to constrain the translation part of the camera extrinsic.

\subsubsection{Lanes}
We optimize lane lines by optimizing the positions of control points. Unlike the point-to-spline residual~\cite{qiao2023online} (For simplicity, it is referred to as PSR in the following text), we propose projected point-to-spline residuals (PPSR) for optimizing control points due to the prediction point error we mentioned earlier. We first determine the four control points $\mathbf{p}_{u_0}, \mathbf{p}_{u_1}, \mathbf{p}_{u_2}, \mathbf{p}_{u_3}$ near the target lane point $\mathbf{L}_j^m$. Then we project $\mathbf{p}_{u_1}$ and $\mathbf{p}_{u_2}$ back onto the image plane to determine the direction vector $\mathbf{d}_{u_2}^{u_1}$. The objective of optimization is to make the lane points detected on the image plane corresponding to $\mathbf{L}_j^m$ as close as to $\mathbf{d}_{u_2}^{u_1}$ (abbreviated as $\mathbf{d}$), namely,
\begin{equation}
\begin{aligned}
&\hat{\mathcal{X}}=\underset{\mathbf{R}_{b}^c, \mathbf{t}_{b}^c, \mathbf{L}_j^m}{\operatorname {arg\,min}} \, \Biggl\{ \sum_{j, m, k} f(j, m, k) \\ 
&\rho ( ||(I - {\mathbf{d}}^{\top}\mathbf{d})(\mathbf{\pi}_{k}^{jm}(p_{u_1}(\mathbf{L}_j^m)) - \mathbf{z}^{jm}_k)||^2_\sigma ) 
+ ||\mathbf{t}_{b}^c - \mathbf{t}_0)||^2_\sigma \Biggl\}.
\end{aligned}
\end{equation}
$p_{u_1}$ is the function for finding $\mathbf{p}_{u_1}$ near the target lane point. There are two advantages of PPSR over PSR. Firstly, the predicted world coordinates of lane points are noisy (whether using IPM or deep learning-based methods), even if we filter them by estimating the error of IPM. PSR will incorporate these errors into the optimization process, while PPSR will not. Secondly, PPSR can unify our optimization of longitudinal markings and lane splines, by which we can optimize map points and IPM matrix simultaneously.

\subsection{Pose Update}
\begin{figure}[!t] 
    \centering
    \subfloat[Port Scenario]{
            \includegraphics[height=2.6cm]{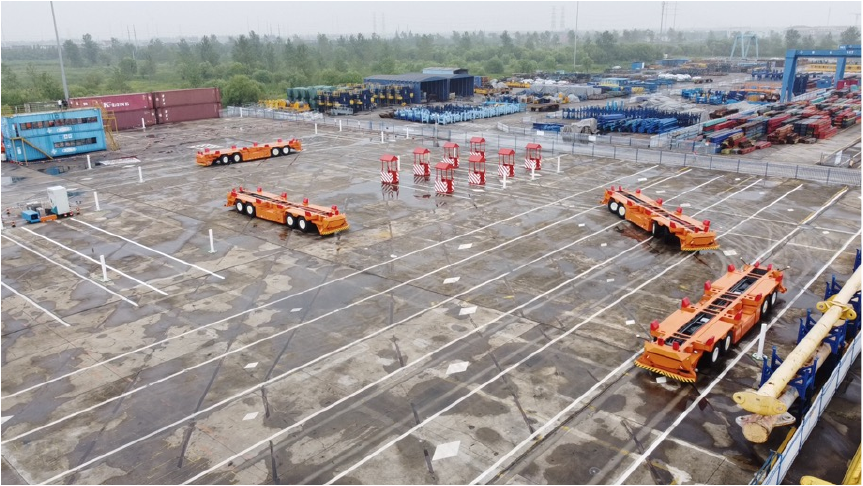}
            \label{subfig: port}
        }  
    \hfill
    \subfloat[Port UGV]{
            \includegraphics[height=2.6cm]{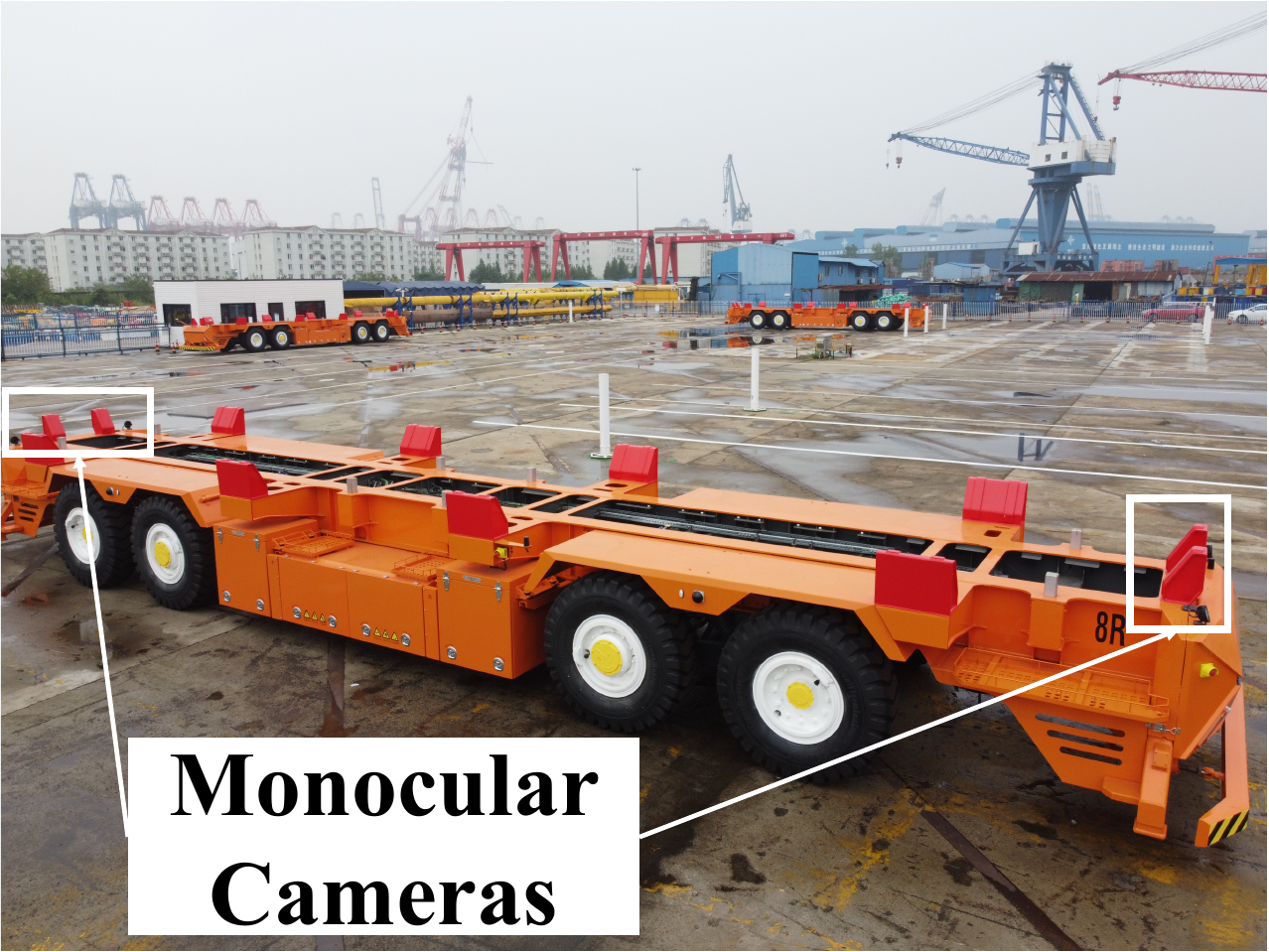}
            \label{subfig: port_igv}
        } \\
    \subfloat[Public Road Scenario]{
        \includegraphics[height=2.6cm]{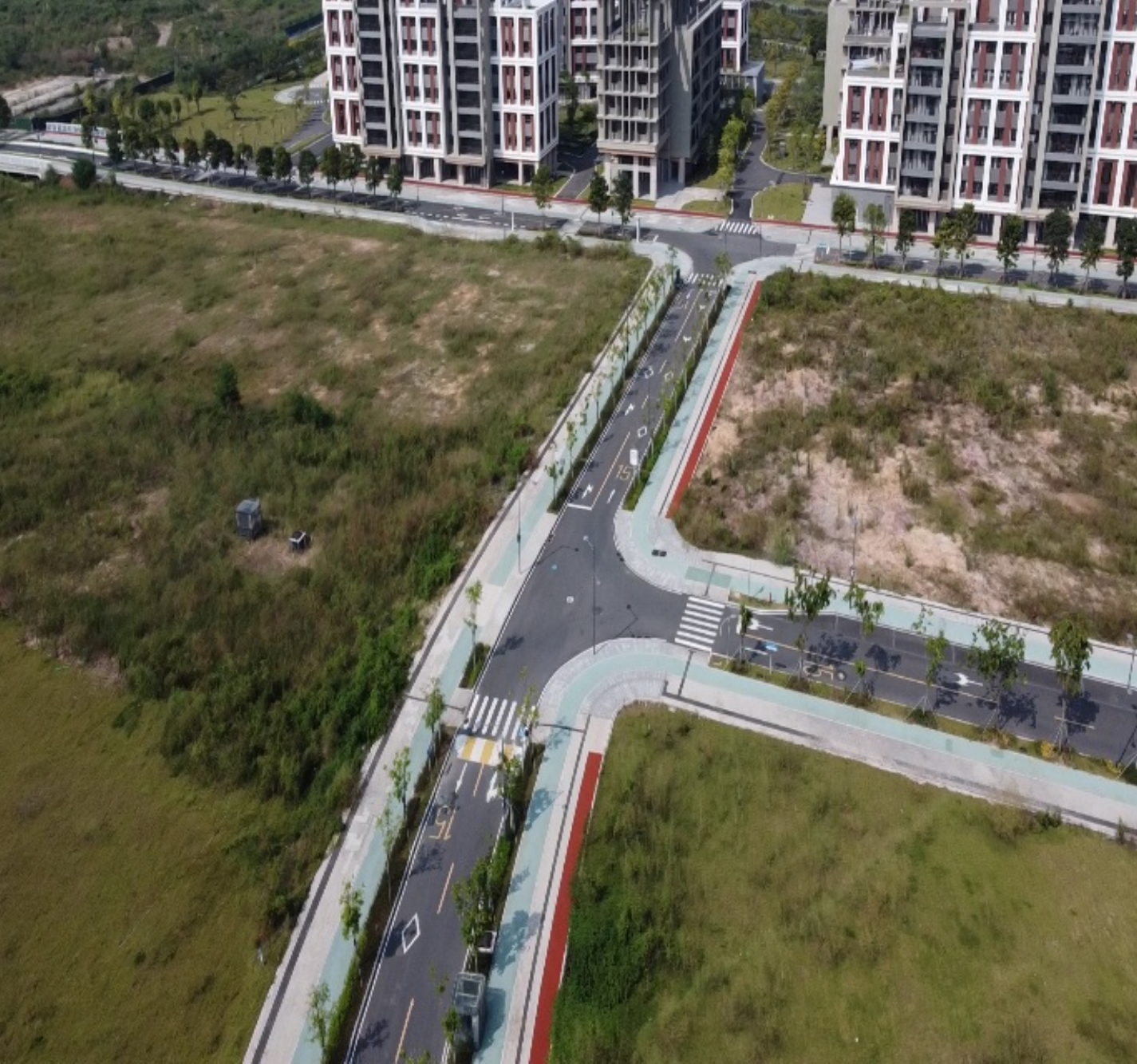}
        \label{subfig: campus}
    }  
    %\hfill
    \subfloat[Public Road UGV]{
        \includegraphics[height=2.6cm]{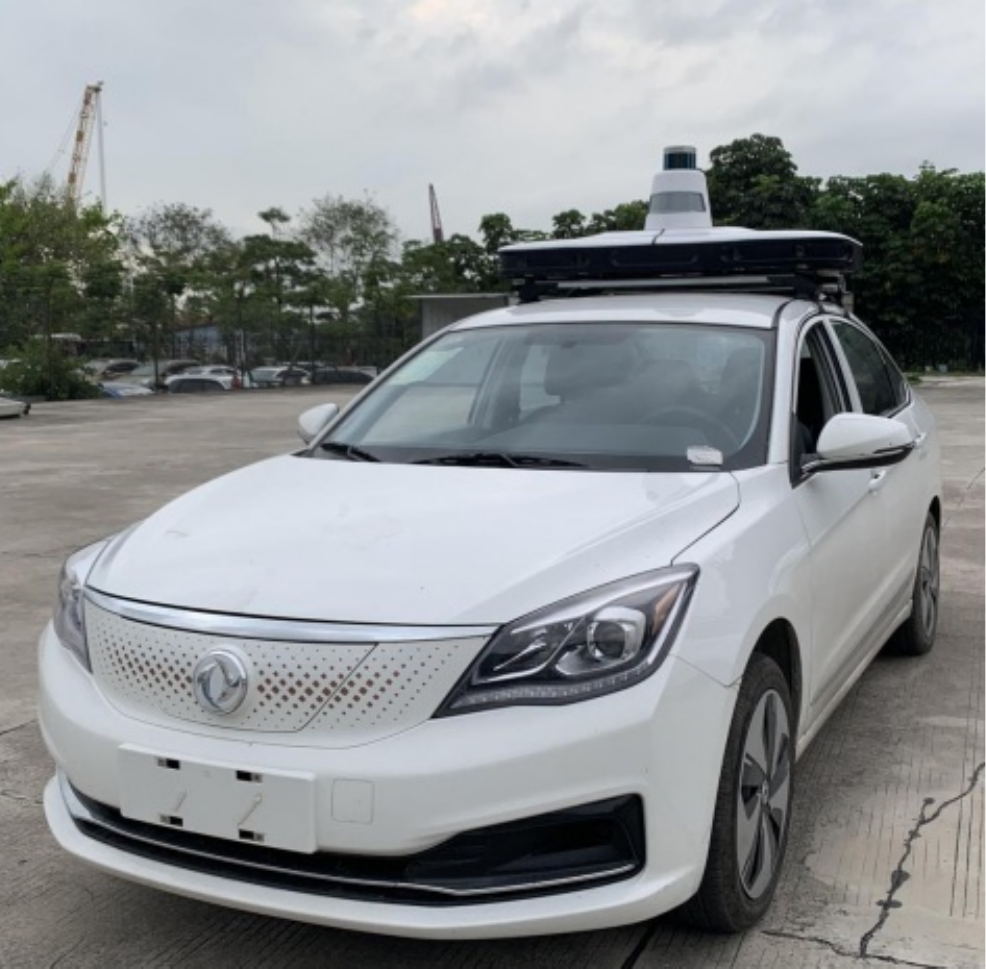}
        \label{subfig: campus_ugv}
    } 
    \subfloat[UGV Camera]{
        \includegraphics[height=2.6cm]{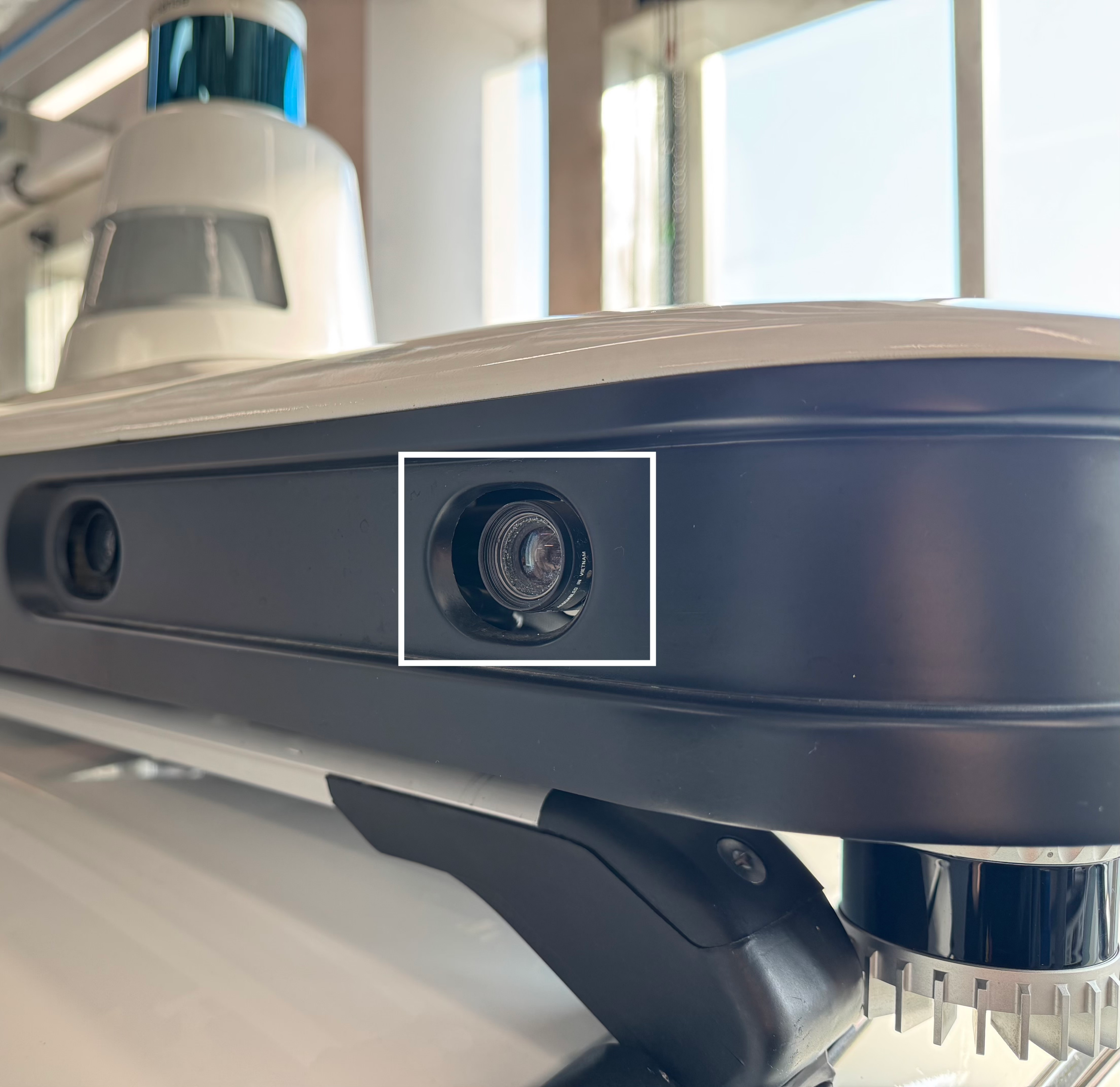}
        \label{subfig: campus_ugv_cam}
    }     
    \\
    \caption{The test scenarios and platforms used in our experiments. For the port scenario, there are hundreds of diamond markings and multiple lane lines on the ground that have been meticulously surveyed. Monocular cameras are installed at the front and rear of the vehicle. The cameras capture RGB 1280 $\times$ 720 images at 20 Hz and are hardware synchronized. For the public road scenario, there are various common traffic ground markings, such as pedestrian crossings, speed limit numbers, straight arrows, etc. A Basler acA2440-20gc monocular camera with a frame rate of 23 FPS installed at the top of UGV is used in our experiment.}
\end{figure}
Since we optimize the position of the camera relative to the vehicle during the map optimization phase, optimizing the pose of the vehicle at the same time may cause conflicts (adjusting the position of the camera relative to the vehicle and the pose of the vehicle have similar effects on the projection results). Therefore, we choose to alternate the map optimization process and pose update process iteratively. The problem of pose update can be formulated as
\begin{equation}
\begin{aligned}
&\hat{\mathbf{T}}_{k}=\underset{\mathbf{T}_{k}}{\operatorname {arg\,min}} \, \Biggl\{ \\&\sum_{i, j, m_1, m_2, k} f(i, m_1, k) \rho (||\mathbf{\pi}_{k}^{i{m_1}}(\mathbf{M}_i^{m_1}) - \mathbf{z}^{i{m_1}}_k||^2_\sigma) \\
&+ f(j, m_2, k) \rho ( ||(I - {\mathbf{d}}^{\top}\mathbf{d})(\mathbf{\pi}_{k}^{jm_2}(p_{u_1}(\mathbf{L}_j^{m_2})) - \mathbf{z}^{jm_2}_k)||^2_\sigma ) \\
&+ ||{(\overline{\mathbf{T}}_k)}^{-1}\mathbf{T}_{k}||^2_\sigma \Biggl\},
\end{aligned}
\end{equation}
where $\overline{\mathbf{T}}_k$ is the prior of the $k$th pose. The effect of pose update is related to the diversity of observation. For example, when driving on a straight road, the lane lines have only lateral positional constraints on pose updates. In such situations, longitudinal markings play a crucial role in supplementing pose updates.

%% file: sections/exp.tex
\section{Experimental Results\label{sec: exp}}

To the best of our knowledge, the existing public HD map data sets cannot provide the global coordinates ground truth for ground markings, lane lines, and IPM homography matrix simultaneously. Therefore, we test our method in two practical scenarios, including an automated port (Fig.~\ref{subfig: port}) where UGVs rely on diamond markings on the ground to complete the visual localization and a public open road (Fig.~\ref{subfig: campus}). Monocular cameras and RTK-GNSS receivers with centimeter-level localization accuracy are equipped on our port UGV (Fig.~\ref{subfig: port_igv}). The public road UGV is
also equipped with a monocular camera (Fig.~\ref{subfig: campus_ugv_cam}) and a PwrPak7/7D-E1 integrated navigation system. To obtain the ground truth, we use Total Station to survey the precise coordinates of the geometric center of the ground markings on the sites manually. For lane lines, we obtain their true positions by manually annotating them in high-precision point cloud maps. The error of ground markings is measured by calculating the Euclidean distance between the position of the markings in the automatically generated map and the ground truth. For the error of lane lines, we follow the evaluation approach in~\cite{qiao2023online}, calculating the Euclidean distance between all generated lane line points and the nearest point in the ground truth as the error.

%%%%%%%%%%%% Table 1  %%%%%%%%%%
\begin{table}[!t] 
\setlength{\abovecaptionskip}{0pt} 
\setlength{\belowcaptionskip}{0pt} 
\renewcommand\arraystretch{1.5} %row height
\renewcommand\tabcolsep{1.5pt}
\centering 
\begin{threeparttable}
    \caption{Map accuracy evaluated using APE. The best results in a sequence are marked in bold. The unit of APE is meter.}
    \label{tab: port map accuracy}
    \begin{tabular}{ c c c c c c c c c c}
    \toprule
    \multicolumn{2}{c}{\multirow{4}{*}{{Baselines}}}& \multicolumn{8}{c}{Sequences} \\ 
    \cmidrule(l){3-10} 
    &&\multicolumn{2}{c}{Sequence 1} & \multicolumn{2}{c}{Sequence 2} & \multicolumn{2}{c}{Sequence 3} & \multicolumn{2}{c}{Sequence 4} \\
    \cmidrule(l){3-4} \cmidrule(l){5-6} \cmidrule(l){7-8} \cmidrule(l){9-10}
    && Marking & Lane & Marking & Lane  & Marking & Lane & Marking & Lane  \\
    \midrule
    IPM & & 0.58 & 0.09 & 0.62 & 0.18 & 0.63 & 0.1 & 0.65 & 0.22 \\
    PGO-IPM & & 0.24 & 0.09 & 0.26 & 0.18 & 0.26 & 0.1 & 0.27 & 0.22 \\
    IPM (+PS) & & 0.24 & 0.07 & 0.26 & 0.17 & 0.26 & 0.08 & 0.27 & 0.19 \\
    \midrule
    Ours (-) & &\textbf{0.16} & 0.2 & \textbf{0.2} & 0.15& \textbf{0.18} & 0.22 & \textbf{0.21} & 0.34 \\
    Ours (+PS) & &\textbf{0.16} & 0.06 & \textbf{0.2} & 0.13 & \textbf{0.18} & 0.07 & \textbf{0.21} & 0.19 \\
    \midrule
    Ours (+PSR) & & \textbf{0.16} & 0.06 & \textbf{0.2} & 0.11 & \textbf{0.18} & 0.07 & \textbf{0.21} & 0.19 \\
    Ours (+PPSR) & & \textbf{0.16} & \textbf{0.05}& \textbf{0.2} & \textbf{0.1}& \textbf{0.18} & \textbf{0.06} & \textbf{0.21} & \textbf{0.16} \\
    
    \bottomrule 
    
    \end{tabular} 
    \begin{tablenotes}
    \RaggedRight
    \item[*] IPM (+PS) means the use of our proposed IPM uncertainty estimation method to filter out the points with high uncertainty. `PS' means `Point Selection'.
    \item[*] Ours (-) means our method uses original IPM points to estimate lane line spline control points.
    \item[*] Ours (+PSR) means our method uses the PSR to optimize the lane line control points. Ours (+PPSR) means our method uses the PPSR to optimize the lane line control points.
    \end{tablenotes}
\end{threeparttable}
\end{table}

\subsection{Port Scenario}
\subsubsection{Evaluation on Map Accuracy}

\begin{figure}[!t] 
    \centering
    \includegraphics[height=5.5cm]{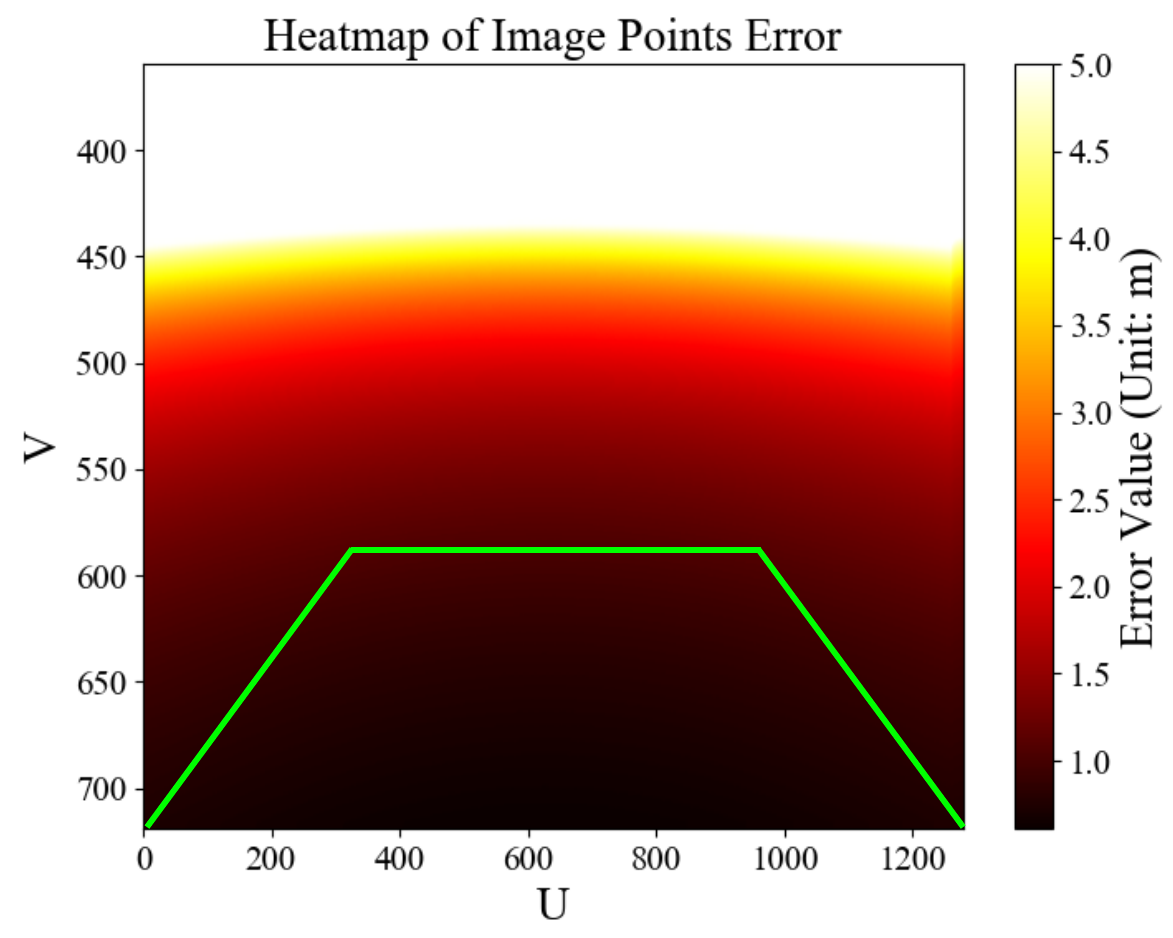}
    \caption{Estimated error of IPM-generated points from a front-view camera image, considering pixel errors in instance segmentation, camera height errors during vehicle motion, and pitch angle errors of the camera relative to the vehicle body frame. The horizontal and vertical coordinates of the figure correspond to the horizontal and vertical coordinates of a real 1280 $\times$ 720 image used in our experiment.}
    \label{fig: ipm error}
\end{figure}
We collected four data sequences in our port scenario, which include multiple lane lines and dozens of ground markings. We evaluated the accuracy of the proposed mapping scheme and compared it with the original IPM method and our previous work PGO-IPM~\cite{10588820} from the perspectives of ground markings and lane lines, respectively.

We calculate the average positional error (APE) for all markings and lane lines involved in each sequence. Table~\ref{tab: port map accuracy} presents the overall evaluation comparison results. It should be noted that the bottom four baselines have the same mapping accuracy for longitudinal markings, as they only differ in lane line mapping. Because of the newly added optimization of the $z$-axis coordinate, our method has further improved the mapping accuracy of the markings compared to PGO-IPM. 

Compared with IPM and Ours (-), IPM (+PS) and Ours (+PS) show better accuracy for lane construction. It indicates that our proposed filtering strategy based on IPM uncertainty estimation can effectively filter out points with large potential errors. We visualize the estimation errors corresponding to each pixel in the image in the form of a heatmap in Fig.~\ref{fig: ipm error}. The pixels we selected for predicting lane control points mostly come from the green box range in the image, with an estimated IPM error of less than 1 m. In general, due to the perspective noise, the farther the scene, the greater the error is~\cite{roadmap}. 

Compared to Ours (+PSR), Ours (+PPSR) demonstrates higher accuracy in lane line mapping, proving that our concerns about PSR are reasonable, and our proposed PPSR is a better optimization strategy in the situation where estimated lane line points may possess errors.
 
\subsubsection{Evaluation on $z$-axis Optimization}

\begin{figure}[!t] 
    \centering
    \includegraphics[width = 0.4\textwidth]{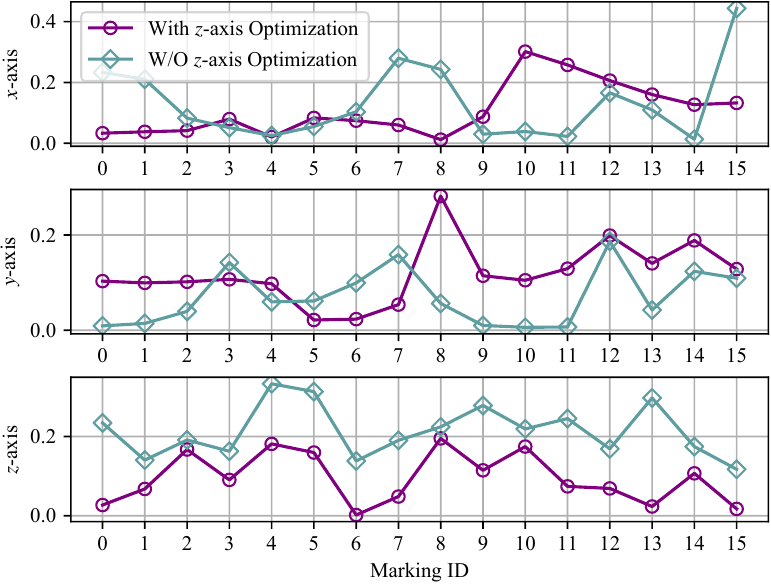}
    \caption{Comparison of marking mapping accuracy for adding $z$-axis coordinate optimization. The horizontal axis indicates the ID of the markings, and the vertical axis shows the positional error of the marking on the three axes respectively, measured in meters.}
    \label{fig: xyz error}
\end{figure}

To further quantify the impact of our optimization on the $z$-axis coordinate, we assessed the marking mapping accuracy with and without $z$-axis coordinate optimization. Fig.~\ref{fig: xyz error} presents a comparison of the accuracy of the $x$-axis, $y$-axis, and $z$-axis coordinates for all markings included in a sequence. The comparison of $z$-axis coordinates error clearly shows that the optimization of the $z$-axis coordinates features a significant improvement, reducing the average error by about 0.1 meters. Notably, adding optimization to the $z$-axis coordinates does not affect the accuracy of the $x$-axis and $y$-axis. In some cases, there was even an improvement in the accuracy of these two axes.

\subsubsection{Evaluation on IPM Homography Matrix Optimization}
\begin{figure}[!t] 
    \centering
    \includegraphics[width = 0.4\textwidth]{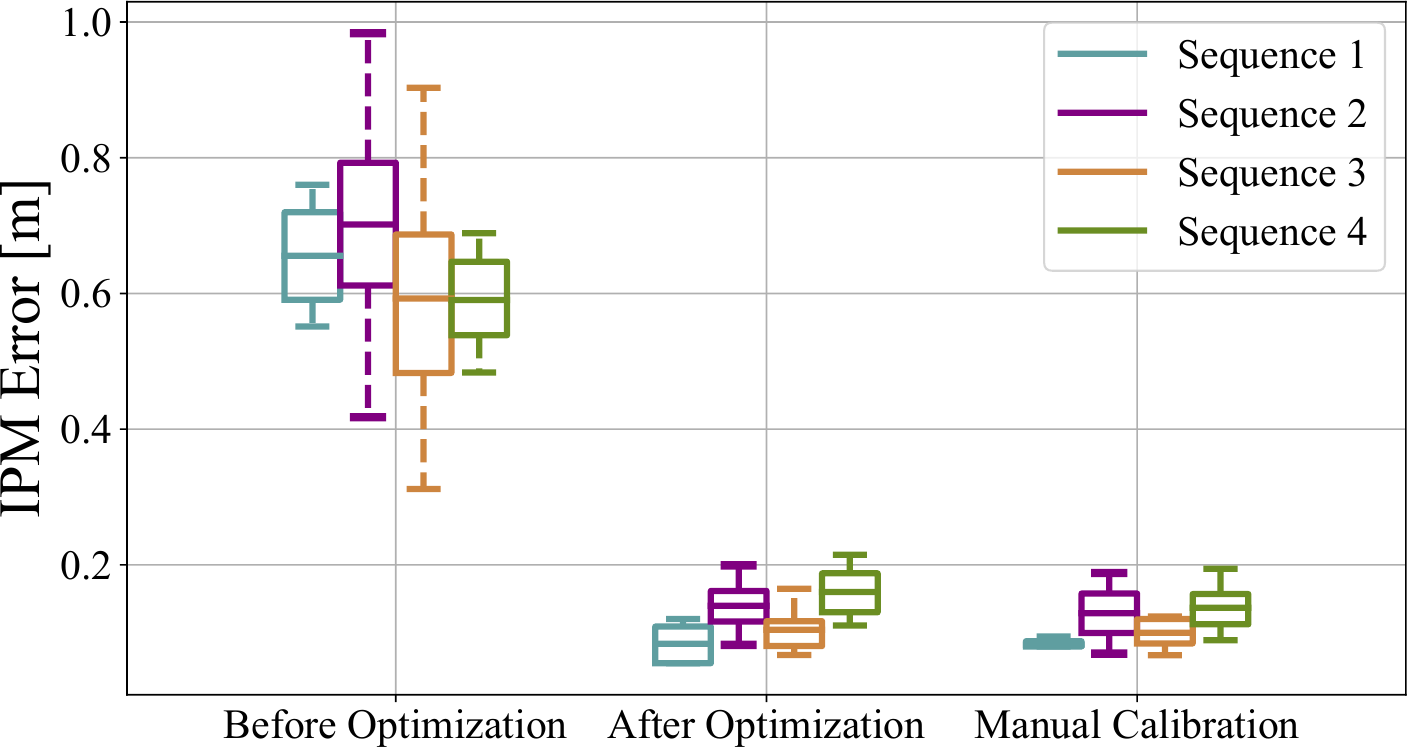}
    \caption{IPM homography matrix error comparison. We use IPM matrices from rough estimation (before optimization), optimized estimation (after optimization), and manual calibration to perform IPM on markings from different data sequences and compare their errors with ground truth.}
    \label{fig: H error}
\end{figure}

In our system, we optimize the IPM homography matrix concurrently with the map optimization. Therefore, in addition to map accuracy, we assessed the optimization effect of the IPM homography matrix separately. To conduct the evaluation, we parked the vehicle in an appropriate position in front of a ground marking. We then used the same pixel coordinates of the markings in the image plane to perform IPM with different homography matrices. Subsequently, we compared the IPM results with the actual position coordinate of the marking. Fig.~\ref{fig: H error} illustrates the accuracy comparison of the IPM homography matrix before and after optimization with calibration results. The data presented in the box plot represents the errors from different corners of the same marking. The results indicates that our optimization process significantly improved the accuracy of the IPM homography matrix and achieved a precision comparable to that of manual calibration.

\subsubsection{Evaluation on Pose Update}
%%%%%%%%%%%% Table 1
\begin{table}[t] 
\setlength{\abovecaptionskip}{0pt} 
\setlength{\belowcaptionskip}{0pt} 
\renewcommand\arraystretch{1.5} %row height
\renewcommand\tabcolsep{5pt} 
\centering 
\begin{threeparttable}
    \caption{Evaluation for the pose update performance of our proposed method. The unit of rotation error is degree and the unit of translation error is meter.}
    \label{tab: pose update}
    \begin{tabular}{c c c c c c c c c}
    \toprule
    \multirow{2}{*}{{Noises}} & \multicolumn{2}{c}{Sequence 1} & \multicolumn{2}{c}{Sequence 2} & \multicolumn{2}{c}{Sequence 3} & \multicolumn{2}{c}{Sequence 4} \\ 
    %\cmidrule(l){3-10} 
    % &&\multicolumn{4}{c}{Port 1} & \multicolumn{4}{c}{Port 2} \\
    % \cmidrule(l){3-6} \cmidrule(l){7-10}
    & Rot & Trans & Rot  & Trans & Rot & Trans & Rot & Trans  \\
    \midrule
    0.1 / 0.1 & 0.04 & 0.03 & 0.05  & 0.01 & 0.03 & 0.02 & 0.02 & 0.02 \\
    0.2 / 0.2 & 0.11 & 0.05 & 0.09 & 0.03 & 0.11 & 0.05 & 0.07 & 0.04 \\
    0.3 / 0.3 & 0.19 & 0.13 & 0.17 & 0.12 & 0.23 & 0.19 & 0.19 & 0.15 \\
    0.4 / 0.4 & 0.3 & 0.27 & 0.33 & 0.29 & 0.32 & 0.26 & 0.31 & 0.24 \\
    0.5 / 0.5 & 0.47 & 0.4 & 0.45 & 0.39 & 0.43 & 0.37 & 0.45 & 0.4 \\
    
    \bottomrule 
    \bottomrule
    \end{tabular} 
\end{threeparttable}
\end{table}
\begin{table*}[!t] 
\setlength{\abovecaptionskip}{0pt} 
\setlength{\belowcaptionskip}{0pt} 
\renewcommand\arraystretch{1.5} %row height
\centering 
\begin{threeparttable}
    \caption{Map accuracy evaluated using APE. The best results in a scenario are marked in bold. The unit of APE is meter.}
    \label{tab: campus_tab}
    \begin{tabularx}{0.95\textwidth}{ >{\centering\arraybackslash}X >{\centering\arraybackslash}X >{\centering\arraybackslash}X >{\centering\arraybackslash}p{14mm} >{\centering\arraybackslash}p{14mm} >{\centering\arraybackslash}p{14mm} >{\centering\arraybackslash}p{14mm} >{\centering\arraybackslash}p{14mm} >{\centering\arraybackslash}p{14mm} >{\centering\arraybackslash}p{14mm} }
    \hline
    \multicolumn{2}{c}{\multirow{3}{*}{Sequences}}& \multirow{3}{*}{\makecell{Marking\\ Type}}& \multicolumn{7}{c}{Baselines} \\ 
    \cmidrule(l){4-10} 
    & & & IPM & PGO-IPM & PersFormer & MLM (-) & MLM (+) & Ours (-) & Ours (+) \\
    \midrule
    \multirow{4}{*}{\makecell{Short \\ ($<$500 m)}} & \multirow{2}{*}{Sequence 1} & Markings & 0.59\,$\pm$\,0.09 & 0.23\,$\pm$\,0.11 & - & - & - & \textbf{0.14}\,$\pm$\,\textbf{0.08} & \textbf{0.14}\,$\pm$\,\textbf{0.08} \\
    & & Lanes & 0.31\,$\pm$\,0.18 & 0.31\,$\pm$\,0.18 & 0.29\,$\pm$\,0.09 & 0.23\,$\pm$\,0.11 & 0.21\,$\pm$\,0.11 & 0.14\,$\pm$\,0.12 & \textbf{0.07}\,$\pm$\,\textbf{0.04} \\

    \cmidrule(l){2-10} 
    & \multirow{2}{*}{Sequence 2} & Markings & 0.56\,$\pm$\,0.1 & 0.24\,$\pm$\,\textbf{0.08} & - & - & - & \textbf{0.17}\,$\pm$\,0.1 & \textbf{0.17}\,$\pm$\,0.1\\

    & & Lanes & 0.37\,$\pm$\,0.28 & 0.37\,$\pm$\,0.28 & 0.18\,$\pm$\,0.11 & 0.21\,$\pm$\,0.11 & 0.20\,$\pm$\,0.12 & 0.16\,$\pm$\,0.13 & \textbf{0.08}\,$\pm$\,\textbf{0.04}\\

    \midrule

    \multirow{4}{*}{\makecell{Long \\ ($>$1000 m)}} & \multirow{2}{*}{Sequence 3} & Markings & 0.62\,$\pm$\,0.11 & 0.28\,$\pm$\,\textbf{0.07} & - &- & - & \textbf{0.22}\,$\pm$\,0.11 & \textbf{0.22}\,$\pm$\,0.11 \\
    & & Lanes & 0.31\,$\pm$\,0.17 & 0.31\,$\pm$\,0.17 & 0.21\,$\pm$\,0.11 & 0.21\,$\pm$\,0.11 & 0.20\,$\pm$\,0.11 & 0.13\,$\pm$\,0.11 & \textbf{0.08}\,$\pm$\,\textbf{0.05} \\

    \cmidrule(l){2-10} 
    
    & \multirow{2}{*}{{Sequence 4}} & Markings & 0.61\,$\pm$\,0.09 & 0.26\,$\pm$\,\textbf{0.08} &- & - & - & \textbf{0.19}\,$\pm$\,0.1 & \textbf{0.19}\,$\pm$\,0.1 \\

    & & Lanes & 0.33\,$\pm$\,0.2 & 0.33\,$\pm$\,0.2 & 0.19\,$\pm$\,0.10 & 0.21\,$\pm$\,0.11 & 0.20\,$\pm$\,0.11 & 0.16\,$\pm$\,0.12 & \textbf{0.09}\,$\pm$\,\textbf{0.06}\\

    \bottomrule 
    \end{tabularx} 
    \begin{tablenotes}
    \RaggedRight
    \item[*] MLM (-) means MLM without lane line optimization. MLM (+) means complete MLM.
    \item[*] Ours (-) means our method without lane line optimization. Ours (+) means the complete version of our method.
    \end{tablenotes}
\end{threeparttable}
\end{table*}

% %%%%%%%%%%%% Table 1
% \begin{table}[t] 
% \setlength{\abovecaptionskip}{0pt} 
% \setlength{\belowcaptionskip}{0pt} 
% \renewcommand\arraystretch{1.5} %row height
% \renewcommand\tabcolsep{5pt} 
% \centering 
% \begin{threeparttable}
%     \caption{Evaluation for the pose update performance of our proposed method. The unit of rotation error is degree and the unit of translation error is meter.}
%     \label{tab: pose update}
%     \begin{tabular}{c c c c c c c c c}
%     \toprule
%     {Noises} & \multicolumn{2}{c}{Sequence 1} & \multicolumn{2}{c}{Sequence 2} & \multicolumn{2}{c}{Sequence 3} & \multicolumn{2}{c}{Sequence 4} \\ 
%     %\cmidrule(l){3-10} 
%     % &&\multicolumn{4}{c}{Port 1} & \multicolumn{4}{c}{Port 2} \\
%     % \cmidrule(l){3-6} \cmidrule(l){7-10}
%     % & Rot & Trans & Rot  & Trans & Rot & Trans & Rot & Trans  \\
%     \midrule
%     0.01 & 0.07 & 0.08 & 0.1  & 0.07 & 0.08 & 0.05 & 0.08 & 0.07 \\
%     0.02 & 0.11 & 0.07 & 0.09 & 0.07 & 0.15 & 0.13 & 0.17 & 0.12 \\
%     0.03 & 0.15 & 0.1 & 0.17 & 0.12 & 0.2 & 0.22 & 0.19 & 0.25 \\
%     0.04 & 0.25 & 0.19 & 0.22 & 0.21 & 0.25 & 0.29 & 0.27 & 0.3 \\
%     0.05 & 0.31 & 0.24 & 0.29 & 0.26 & 0.35 & 0.37 & 0.33 & 0.35 \\
    
%     \bottomrule 
%     \bottomrule
%     \end{tabular} 
% \end{threeparttable}
% \end{table}

In order to quantify the effectiveness of our method on pose updates, we randomly added noise to the original pose data to simulate possible errors in reality. Table~\ref{tab: pose update} demonstrates the pose update effect of our system. The first column of the table represents the noise level, e.g., 0.1 / 0.1 represents we added random errors with mean values of 0.1 degrees and 0.1 meters for rotation (yaw) and translation part of the pose, respectively. From the evaluation results, it can be seen that our system demonstrates an ability to effectively reduce pose errors under varying levels of noise influence. Notably, the reduction in translation errors is more pronounced than that of rotation errors. We attribute this difference to the relatively uniform observation angle of the ground markings during testing. If multiple observation angles are available for the same markings, it could lead to a further reduction in rotation errors.

% Additionally, we observed that when noise levels reach a certain level, e.g., 0.5 / 0.5, the extent of error correction diminishes significantly. This suggests that our method has a limit to its pose correction capabilities. We believe this limitation arises from the interdependence of the mapping and pose correction processes. When noise errors are excessively high, the mapping error cannot be effectively minimized through the optimization process, which in turn hampers pose correction efficacy.

\subsection{Public Road Scenario}
\begin{figure}[!t] 
    \centering
    \includegraphics[width = 0.475\textwidth]{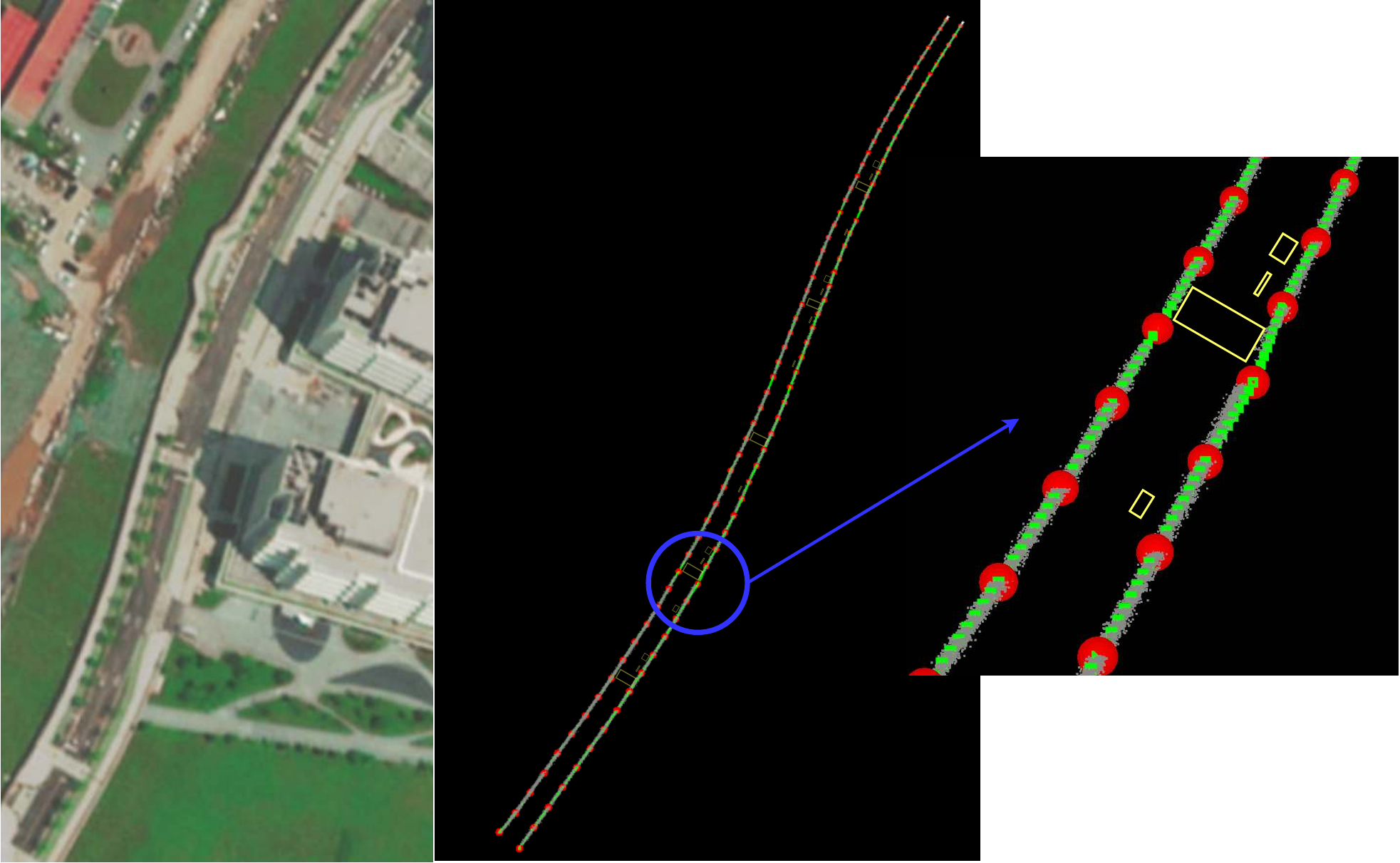}
    \caption{Road mapping example on a public road. The left image is a satellite-captured image of the road. The lane lines on the road surface are presented in the form of Catmull-Rom spline. The red spheres represent control points, and the green spheres represent points sampled on the spline. Other ground markings are uniformly represented by yellow quadrilateral bounding boxes.}
    \label{fig: road mapping}
\end{figure}

In order to conduct more comprehensive testing, we conducted tests on a public road that is a common autonomous driving environment. We collected four distinct data sequences on the road. These sequences were recorded across various road segments, encompassing routes of different lengths. Fig.~\ref{fig: road mapping} shows an example of road mapping for one section of the road. As we introduced in Section~\ref{sec: meth}, we represent the longitudinal markings in the form of polygons, and the lane lines are depicted using Catmull-Rom splines.

\subsubsection{Evaluation on Map Accuracy}

In this experiment, we compared our method not only with IPM and PGO-IPM but also with the state-of-the-art (SOTA) lane mapping framework PersFormer~\cite{10.1007/978-3-031-19839-7_32} and MonoLaneMapping (MLM)~\cite{qiao2023online}. PersFormer models lane lines as points, while MLM uses splines. We assessed the mapping accuracy of these baselines across four data sequences. The overall evaluation results are summarized in Table~\ref{tab: campus_tab}. The data presented in the table includes the mean and standard deviation, indicated by addition and subtraction signs. For longitudinal markings, our results align closely with those observed in the port scenario. Our method significantly reduces the three-dimensional coordinate error of longitudinal markings when compared to IPM and PGO-IPM, achieving accuracy near the centimeter level. 

\begin{figure*}[!t] 
    \centering
    {
        \includegraphics[width=0.31\textwidth]{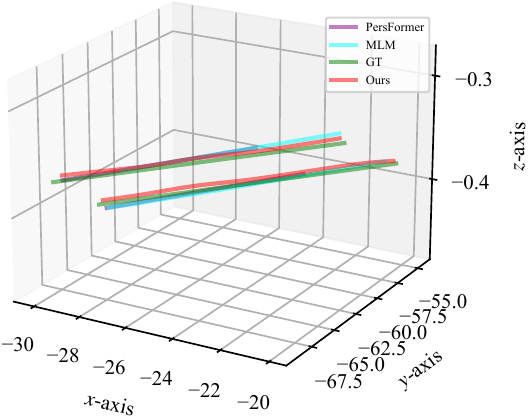}
        \label{fig: lane1}
    }
    {
        \includegraphics[width=0.31\textwidth]{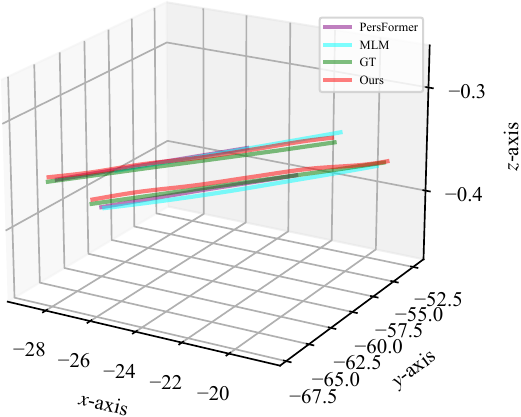}
        \label{fig: lane2}
    }
    {
        \includegraphics[width=0.31\textwidth]{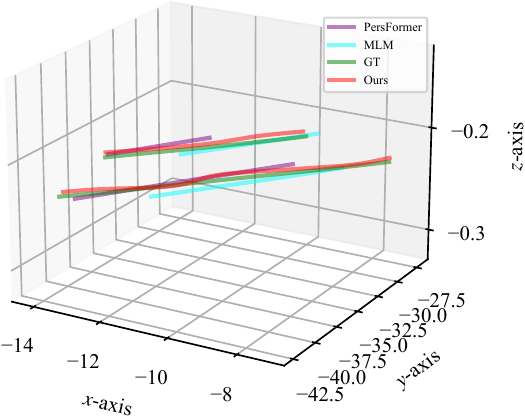}
        \label{fig: lane3}
    }
    \centering
    {
        \includegraphics[width=0.31\textwidth]{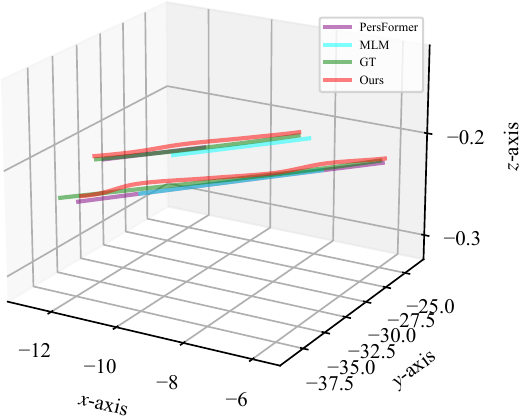}
        \label{fig: lane4}
    }
    {
        \includegraphics[width=0.31\textwidth]{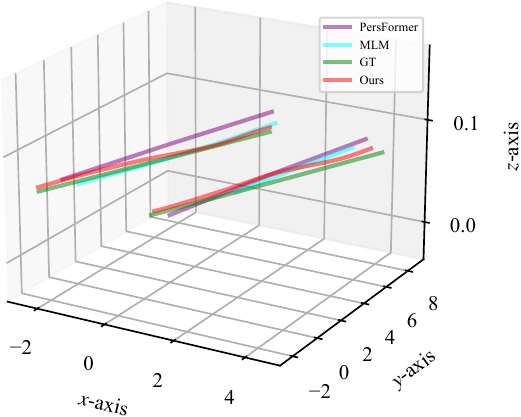}
        \label{fig: lane5}
    }
    {
        \includegraphics[width=0.31\textwidth]{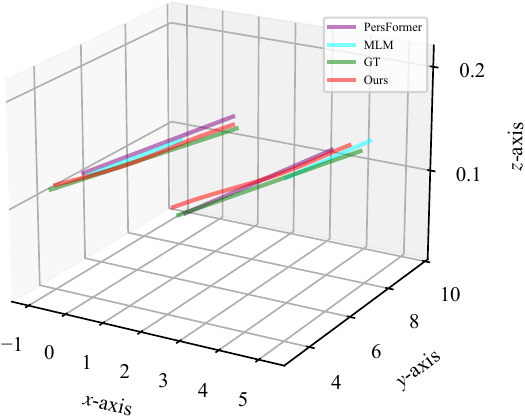}
        \label{fig: lane6}
    }
    \caption{Comparison of lane line mapping effects. The figures show the discrepancies between the predicted 3D lane line points using different methods and ground truth across various road sections. Our approach generally predicts lane lines that closely match the length of the ground truth. In contrast, other methods tend to predict only a portion of the actual lane line length. Additionally, our predicted lane lines typically exhibit lower lateral error compared to other baselines.}
    % \vspace{-15pt}
    \label{fig: map error}
\end{figure*}

In terms of lane line mapping, our approach outperformed both PersFormer and MLM, exhibiting a lower standard deviation. Our approach reduced their error by approximately 50\%, reaching centimeter-level accuracy. Fig.~\ref{fig: map error} illustrates the lane mapping performance across several segments of public roads. This comparison shows that the positions of lane lines mapped by our method are generally closer to the ground truth than those produced by the other two SOTA methods. 

To evaluate our point selection strategy, we specifically compared our method with MLM in the configuration without the optimization process. Under this setting, there are two key differences between our method and MLM. Firstly, our method estimates control points based on predicted points from IPM, while MLM relies on predictions from PersFormer. Secondly, during control point estimation, our approach will filter out points with potentially large errors by analyzing the errors from the IPM process. The experimental results indicate that despite IPM's lower accuracy compared to PersFormer, our method still achieves higher control point prediction accuracy. This underscores the validity and importance of our strategy for analyzing IPM errors and selecting high-quality original lane line points. 

\subsubsection{Evaluation on Lane Optimization Effect}

\begin{table}[!t] 
\setlength{\abovecaptionskip}{0pt} 
\setlength{\belowcaptionskip}{0pt} 
\renewcommand\arraystretch{1.5} %row height
\centering 
\begin{threeparttable}
    \caption{Comparison of map accuracy improvement by optimization process. The data unit in the table is meter.}
    \label{tab: coe}
    \begin{tabular}{ c  c  c  c  c }
    \toprule
     & \makecell{Sequence 1} & \makecell{Sequence 2} & \makecell{Sequence 3} & \makecell{Sequence 4} \\ 
    \midrule
    {MLM} & 0.02 & 0.01 & 0.01 & 0.01  \\
    \midrule
    {Ours} & \textbf{0.08} & \textbf{0.09} & \textbf{0.07} & \textbf{0.07}   \\
    \bottomrule 
    \end{tabular} 
    \vspace{-1.5em}
\end{threeparttable}
\end{table}

Table~\ref{tab: coe} presents the accuracy improvements in lane line mapping achieved by our method and MLM after adding the optimization process. The data in the table indicate the amount of reduction in error. The data demonstrates that our PPSR optimization method outperforms the PSR strategy of MLM in enhancing lane mapping accuracy. As previously mentioned, when errors exist in the predicted lane line points, there is a limit to the accuracy that can be achieved in optimizing lane control points for PSR strategy. On the contrary, utilizing projection techniques can mitigate the impact of errors in predicted points. 

\subsubsection{Evaluation on Data Volume}

\begin{table}[!t] 
\setlength{\abovecaptionskip}{0pt} 
\setlength{\belowcaptionskip}{0pt} 
\renewcommand\arraystretch{1.5} %row height
\centering 
\begin{threeparttable}
    \caption{Comparison of map storage and data volume of lane line map constructed by different baselines.}
    \label{tab: msdv}
    \begin{tabular}{ c  c  c  c  }
    \toprule
     & \makecell{Map Distance} & \makecell{Map Storage} & \makecell{Point Number} \\ 
    \midrule
    \multirow{2}{*}{{PersFormer}} & 300 m & 150 KB & 47418  \\
    & 1360 m & 13 MB & 625156 \\
    \midrule
    \multirow{2}{*}{{PGO-IPM}} & 300 m & 251 KB & 84642  \\
    & 1360 m & 7 MB & 383710 \\
    \midrule
    \multirow{2}{*}{{Ours}} & 300 m & \textbf{2 KB} & \textbf{96}   \\
     & 1360 m & \textbf{9 KB} & \textbf{453} \\
    \bottomrule 
    \end{tabular} 
    \vspace{-1.5em}
\end{threeparttable}
\end{table}
The primary purpose of utilizing Catmull-Rom splines for modeling lane lines, rather than individual lane line points, is to optimize data storage and compress maps. Table~\ref{tab: msdv} presents a comparison of the data volume and storage space required by lane line maps created using PersFormer, IPM, and our method. The results clearly indicate that our approach significantly reduces both data volume and storage space. By employing splines to represent lane lines, we only need to store the control points, rather than all the individual lane line points. This modeling technique effectively contributes to the downsampling of lane line data.

%% file: sections/conclusion.tex
\section{Conclusion}
In this work, we proposed a low-cost and unified framework for automatic vectorized road mapping based on enhanced IPM. In the framework, ground markings are depicted using polygons uniformly and lane lines are represented by Catmull-Rom splines. The map elements, IPM homography matrix, and vehicle poses are jointly optimized under the guidance of instance segmentation results. Considering the dependence of IPM on the assumption of coplanarity and the pose errors that may exist during the mapping process, we introduced optimization on the $z$-axis coordinates of map points and correction process for the vehicle poses. The test in multiple practical scenarios demonstrated that our mapping method greatly improved the accuracy of road mapping compared to IPM and our preliminary work PGO-IPM. The optimization of $z$-axis coordinates of map points effectively further improved the map precision. The pose update process that alternated with map optimization could effectively reduce the underlying errors in vehicle poses. The accuracy of the optimized IPM homography matrix is similar to that of manual calibration. In terms of lane line mapping, our proposed method presents better accuracy compared to the SOTA works and reaches centimeter-level accuracy. Our future work includes estimating the confidence level of map elements and updating changes to them. Our goal is to automate the generation and updating of road maps, facilitating high-frequency updates of road maps in the form of crowdsourcing in the future.